\documentclass[journal]{IEEEtran}

\usepackage[utf8]{inputenc} 
\usepackage{hyperref}       
\usepackage{url}            
\usepackage{booktabs}       
\usepackage{amsfonts}       
\usepackage{nicefrac}       
\usepackage{microtype}      
\usepackage{lipsum}
\usepackage{graphicx} 
\graphicspath{ {./images/} }
\usepackage[margins]{trackchanges}
\usepackage{booktabs}
\usepackage{bbm}
\usepackage{amssymb}
\usepackage{float}
\usepackage{caption}
\usepackage[export]{adjustbox}
\usepackage{amsmath}
\usepackage{nccmath}
\usepackage{balance}
\usepackage{listings}
\usepackage[dvipsnames]{xcolor}

\title{Multi-scale Cloud Detection in Remote Sensing Images using a Dual Convolutional Neural Network}

\author{Markku~Luotamo$^*$,~
        Sari~Metsämäki~
        and~Arto~Klami
\thanks{$^*$M.Luotamo and A.Klami: University of Helsinki, Dept. of Computer Science}%
\thanks{Sari Metsämäki: Finnish Environment Institute}}

\addeditor{AK}
\addeditor{ML}
\addeditor{SM}

\usepackage{hyperref}
\hypersetup{
    colorlinks=true,
    linkcolor=blue,
    filecolor=magenta,      
    urlcolor=cyan,
}
 
\begin{document}
\maketitle

\begin{abstract}
Semantic segmentation by convolutional neural networks (CNN) has advanced the state of the art in pixel-level classification of remote sensing images. However, processing large images typically requires analyzing the image in small patches, and hence features that
have large spatial extent still cause challenges in tasks such as cloud masking. To support a wider scale of spatial features while simultaneously reducing computational requirements for large satellite images, we propose an architecture of two cascaded CNN model components successively processing undersampled and full resolution images. The first component distinguishes between patches in the inner cloud area from patches at the cloud's boundary region. For the cloud-ambiguous edge patches requiring further segmentation, the framework then delegates computation to a fine-grained model component. We apply the architecture to a cloud detection dataset of complete Sentinel-2 multispectral images, approximately annotated for minimal false negatives in a land use application. On this specific task and data, we achieve a 16\% relative improvement in pixel accuracy over a CNN baseline based on patching. 

\end{abstract}

\begin{figure*}[t!]
  \includegraphics[width=0.8\textwidth, center, trim=0 2.5cm 3cm 0]{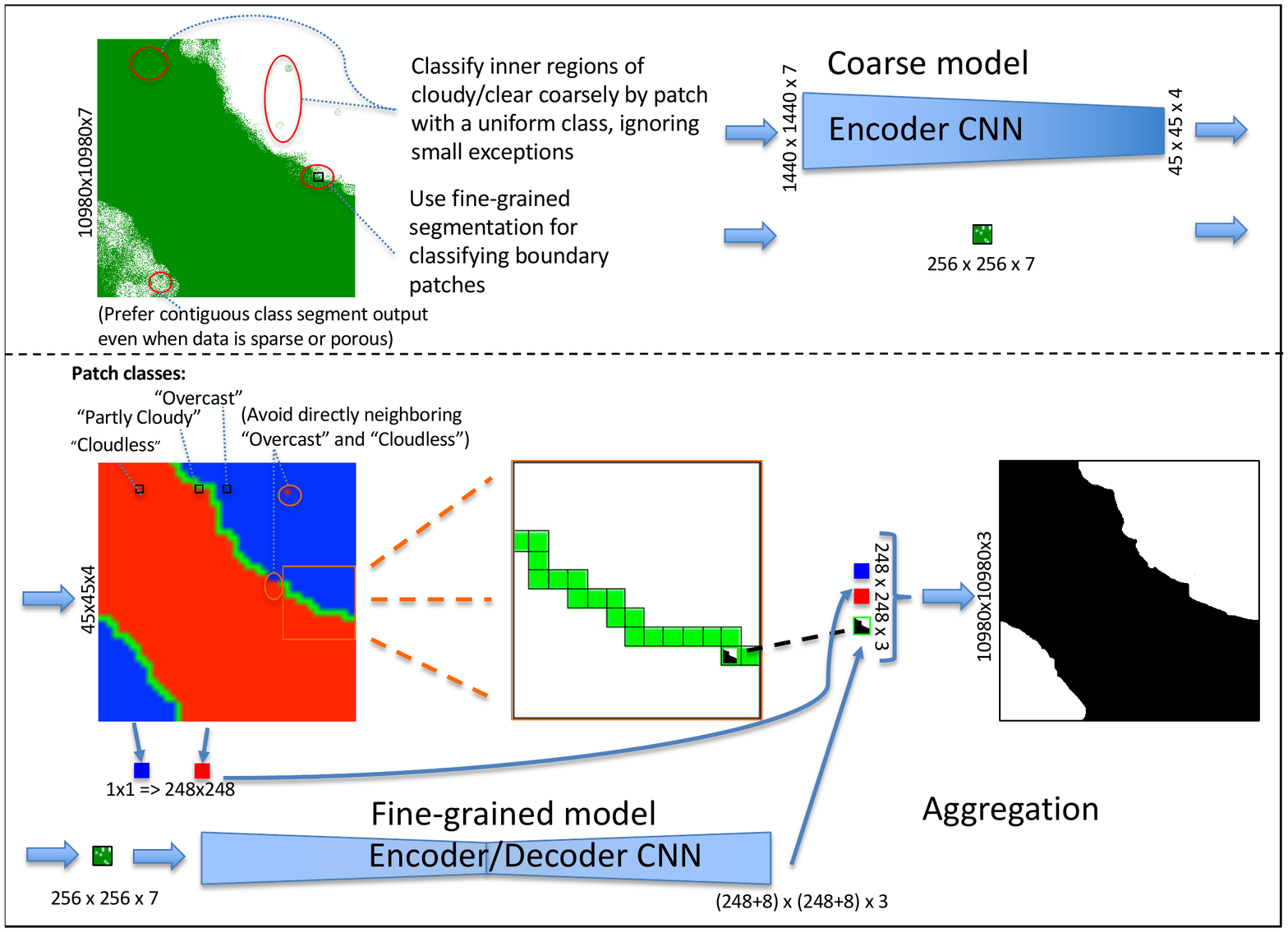} %
  \caption{
  Illustration of the main concepts and the overall workflow. \textbf{Top:} The \textit{coarse} component analyzes complete MSI images to classify patches as overcast, cloudless or partly cloudy (at the boundary of the cloudy and clear areas). \textbf{Bottom:} Subsequently, the \textit{fine-grained} component classifies pixels of "partly cloudy" patches at full resolution.
  }
  \label{graphstract}
\end{figure*}

\section{Introduction}
\label{sec:intro}

\IEEEPARstart{R}{emote}
  sensing analytics applications, such as classification of land use, vegetation, urban structures  or crop type \cite{marmanis2016semantic,langkvist2016classification,kussul2017deep}, make use of semantic segmentation i.e. pixel-level classification of images in detecting and visualizing shapes of phenomena and objects on aerial and satellite imagery. When ground-level feature segmentation is based on optical satellite imagery and thus ground reflectance, presence of atmospheric clouds or haze in images is inevitable. 
  
  There are various types of clouds as to their shape, optical thickness, extent, height etc. A cloud may affect only one individual pixel, may form vast contiguous surfaces or may be fragmented so that it contains small gaps where visibility to the earth is not completely obscured. Shadows cast by clouds are also problematic \cite{mahajan2019cloud}.  In optical remote sensing, the main problem emerging from presence of clouds is that they partly or totally block the view from the satellite sensor to the ground target. If unaccounted for, clouds and haze can result in false interpretation of an image and therefore cause incorrect conclusions from the application's perspective.  To identify the usable pixels of an image, automatically generated \textit{cloud masks}  \cite{baetens2019validation, ackerman2015modis} that themselves are semantic segmentations, are invariably used for optical satellite image applications interpreting ground-level phenomena. For interpretation within a given area of interest, typically a representation of cloudless optical reflectance is required. Depending on the extent of cloud cover in the available imagery, this may call for one or more images, each filtered by its respective cloud mask. When several images are used, their cloudless pixels are composed into a cloudless mosaic image.
 
 Early and still often-used pixel-level classification and estimation methods were based on computationally simple engineered features involving band arithmetics, calculated index values, thresholding or decision trees  \cite{kamavisdar2013survey}. More recently, machine learning (ML) methods such as the Support Vector Machine (SVM) \cite{li2015cloud},  
 Markov Random Fields \cite{le2009use}, and in particular deep learning and convolutional neural networks (CNN) have been used for cloud detection
 \cite{mateo2017convolutional}.
 Today, models based on state-of-the-art CNN architectures, such as fully convolutional networks (FCN)  \cite{long2015fully}, UNet \cite{ronneberger2015u} or their derivatives, originally developed
 for semantic segmentation of RGB photographs and biomedical images, are increasingly being used also for remote sensing \cite{dronner2018fast, mateo2017convolutional, shi2016cloud, xie2017multilevel}.

Multispectral satellite image size runs in a magnitude markedly different from ordinary photographs. For example, a Sentinel-2 multispectral instrument (MSI) image consists of 13 spectral bands in resolutions of 10m (10980x10980 = 121 Mpx) to 60 m (1830x1830 px). With all bands resampled, for example, to the most accurate 10m resolution, an uncompressed image would consume gigabytes of memory. Even though many CNN-based segmentation methods theoretically scale to arbitrary image sizes, and the processing power of accelerator units such as GPUs is continuously increasing, accelerator memory remains a practical limit \cite{chen2016training,wang2018superneurons}. The problem becomes all the more pronounced with high-resolution multispectral satellite images; loading even a single complete full-resolution Sentinel-2 image and processing it with a deep CNN requires more memory than what is available on any single GPU,
not to speak of a mini-batch of full images.
Solutions to this problem usually involve
patching \cite{shao2019cloud, yang2019cdnet, mateo2017convolutional} or undersampling \cite{dronner2018fast}. Patching partitions the image into small sub-images and analyzes each one separately, which solves the memory issue but results in loss of global information and limits detection of spatially wide features, since information outside the current patch has no influence on the segmentation. In turn, undersampling  naturally loses local information and does not provide pixel-level segmentation at the original resolution. To manage either type of information loss and to detect features of large spatial extent, such as atmospheric clouds, we propose a novel practical segmentation architecture that improves retention of both global and local information. 

Our architecture consists of two cascaded CNN model components successively processing undersampled and full resolution images, as illustrated in Figure~\ref{graphstract}. The first \textit{coarse} component is a modified CNN image classifier that receives an \textit{undersampled} input of a complete image and classifies a full set of fixed-size patches of the whole image with a single class for each patch, in a single pass. To accommodate maximal input resolution with limited memory, we keep the coarse model compact by not including a decoder component, but truncating a standard classifier architecture just before its final encoding layer. The cells of the resulting sparse grid correspond to patches subdivided from the original image. The coarse model assigns one of four classes to each patch. The "Partly Cloudy" class signifies cloud ambiguity and hence a need for pixel-level classification within the patch, whereas the other classes assign a single uniform class to all pixels of the patch  (Cloud/Clear/No Data). 
If a patch is classified as "Partly Cloudy",
only then is a second \textit{fine-grained} component used for pixel classification within the patch. This second model is a conventional CNN encoder-decoder that classifies each pixel of its input at the \textit{original full resolution}, using an encoder backbone and a decoder. The two-component coarse-fine architecture enables efficient semantic segmentation of arbitrarily large images while retaining more global and local information than would be possible using exclusively patched or undersampled inputs on a single CNN encoder-decoder.

The requirements of this cascaded architecture are flexible and allow a wide range of choices in assigning different specific CNN architectures as bases for its respective coarse and fine-grained subnetworks. For the fine-grained component, we evaluate a set of recent CNN encoder-decoder architecture variants for semantic segmentation, including PSPNet \cite{zhao2017pyramid}, UNet \cite{ronneberger2015u}, FPN \cite{lin2017feature} and Linknet \cite{chaurasia2017linknet}. We vary the encoder part of these CNNs, choosing from different baseline and state-of-the-art encoder/classifier architectures, including VGG-16 \cite{simonyan2014very}, ResNet-50 \cite{he2016deep}, SEResNeXt-50 \cite{hu2018squeeze}, EfficientNet \cite{tan2019efficientnet} and Inception-v3 \cite{szegedy2016rethinking}. From this set of encoding classifier architectures, we also select the best-performing ones for evaluation as a basis for the coarse classifier.

We train and evaluate the models on a reference dataset of Sentinel-2 images and cloud masks annotated originally for a land use application \cite{corinelc}. The masks had been manually extended from automated masks to ensure non-cloud-contaminated pixels and to avoid model shortcomings described e.g. in \cite{cloudmask_assess}. 
The annotation guidelines instructed towards
\textit{contiguous} cloud regions, as opposed to e.g. porous masks with an abundance of small holes. Small cloud-free areas or pixels had routinely been discarded during annotation. In order to reproduce masks of this nature, we promote high \textit{recall} (see Section \ref{sect_metrics} on metrics) of cloud pixels within inner cloud regions, allowing a small decrease in precision as a tradeoff. This approach serves many applications better than high precision for a cloud class, which would leave part of the cloudy pixels interpreted as cloud-free. Besides training the model on cloud masks annotated with high recall, we also promote cloud recall and accurate detection of the border region between cloudy and cloudless areas by designing a loss function specifically for these purposes.

To summarize, the main contributions of this work are:
\begin{enumerate}
    \item A novel semantic segmentation method for high-resolution multi-spectral images using dual cascaded convolutional neural networks. An efficient coarse segmentation retains global patterns and is further focused using a fine-grained model to full resolution at narrow regions of annotation borders, requiring fine-grained processing only locally.
    \item A loss function to identify the border segments for fine-grained segmentation, to compensate for the fact that the border areas have a naturally low proportion in the sample distribution, and to explicitly favor contiguous segments for improved emulation of manually dilated cloud masks of the reference data.
    \item Demonstration of the approach on cloud segmentation of MSI image data, outperforming both state-of-the-art standalone CNN architectures and well-known baseline cloud detection models in reproducing the high-recall annotations.
\end{enumerate}

\section{Related work}
\label{sec:related}

Optical remote sensing analytics has traditionally used cloud detection methods of thresholding, band arithmetics or feature engineering, see for instance work by Ackerman et al. on MODIS data \cite{ackerman1998discriminating, ackerman2015modis}, or earlier works by Stowe or Cihlar et al. \cite{stowe1988nimbus}\cite{cihlar1994detection}. Since they are computationally efficient to implement, evolved versions of e.g. thresholding decision trees continue to be practical tools today applied to more recent generations of optical satellite imagery for cloud detection \cite{irish2006characterization,zhu2015improvement} as well as other types of pixel-level classification, e.g.  snow masks by Metsämäki et al. \cite{metsamaki2015introduction}  used in the EU/Copernicus Global snow monitoring service. Foga et al. \cite{foga2017cloud} evaluate several cloud detection algorithms against a Landsat validation mask and take preference for the thresholding-based CFMask due to its global applicability and no need for retraining as opposed to machine learning methods.

Although there were pioneer efforts to apply ML and even neural networks \cite{visa1991cloud} to cloud detection, developments in computing power and increased accuracy of new algorithms have given ground to increased use of various machine learning methods.
In particular, advances in computer vision have inspired supervised convolutional neural networks to be applied to semantic segmentation also in the remote sensing and cloud detection context. For example, Mateo et al.  \cite{mateo2017convolutional} showed that their CNN outperformed both a gradient boosting machine and a fully connected multi-layer perceptron, even when the latter two were provided additional features besides the band data.

One of the main challenges in adopting semantic image segmentation advances in remote sensing has been the high dimensionality of the imagery. This still remains a practical challenge despite growing literature, and typical neural network approaches still either train exclusively on small full-resolution patches or on heavily undersampled images.
In the context of cloud masking, Shao et al.\cite{shao2019cloud} use a CNN for segmenting inputs of 128x128x10 MSI patches,
Yang et al. \cite{yang2019cdnet} apply a CNN for 321x321 RGB or grayscale images obtained by patching a downsampled MSI image, 
and 
Moharejani et al. \cite{mohajerani2018cloud} used patches of 196x196x4 in combination with QA snow/ice masks.
A CNN encoder-decoder of Segal-Rozenheimer et al. \cite{segal2020cloud}, inspired by DeepLab \cite{chen2017deeplab}, uses a module of varying-size dilated convolutions before the feature extraction layer, and eventually trains the network on 256x256 patches.

Besides patching and downsampling, large image size can be addressed by 
generating superpixels  \cite{achanta2012slic} i.e. clusters of similar and adjacent pixels, and then classifying parts of the image only at the superpixel level.
Shi et al. \cite{shi2016cloud} assigned a cloud probability to each superpixel's center pixel, based a CNN-classified image patch extracted to center at the same pixel. Xie et al. \cite{xie2017multilevel} followed a similar procedure, but using patches of two different resolutions for determining the cloud status,
and Liu et al. \cite{liu2018super} used CNNs and deep forests on pre-computed cloud superpixels.

Other remote sensing segmentation applications have benefited from the use of CNNs as well, for instance land cover and crop classification \cite{kussul2017deep}. We are inspired by the properties of Fully Convolutional Networks and the UNet architecture and their descendants, as were other authors \cite{mohajerani2018cloud, yang2019cdnet,dronner2018fast} that used them recently on remote sensing data. However, we apply CNNs in a setting of approximate mask annotations to an MSI image at full and reduced resolutions. The closest work is that of Miyamoto et al. \cite{miyamoto2018object}, who applied a two-step convolutional network to remote sensing object detection, optimizing for recall. However, their application is far removed from cloud detection and classifies patches instead of pixels.
Our approach also has the advantage of modularity; we can use various architectures as building blocks of the cascaded solution and assume the choices that provide the best overall accuracy, as demonstrated in Section~\ref{ch_results}.

Our work also relates more generally to machine learning research on semantic segmentation with various forms of approximate or weak supervision, developed to reduce the high cost of pixel-level annotation of large images.
The most common form of weak annotation is to consider bounding boxes surrounding the objects \cite{papandreou2015weakly}, 
but more elaborate approaches are also being studied. For example, multiple instance learning (MIL) strategies where an annotation signifies that an object is to be found somewhere within the indicated area has been used for segmentation of medical images \cite{kraus2016classifying}, whereas Shen et al. developed a method that can be trained on crude annotations, each marked somewhere within the object \cite{shenscribble}.
Pathak et al. \cite{pathak2014fully}, in turn, concentrate segmentation around a single maximum-probability pixel. We consider annotations that are supersets of the positive instances and hence formally fit within the MIL framework. However, MIL algorithms are typically developed for scenarios where the positive class covers only a small fraction of the indicated area, whereas in our case, a majority of the pixels annotated as cloudy are indeed cloudy. Hence, the property is better addressed by an improved loss function (Section~\ref{lossrationale}) instead of dedicated MIL algorithms. 

\section{Method}
\label{method_ch}

\begin{figure*}[t]
  \includegraphics[width=\textwidth, trim=0 13cm 3cm 0, clip, center]{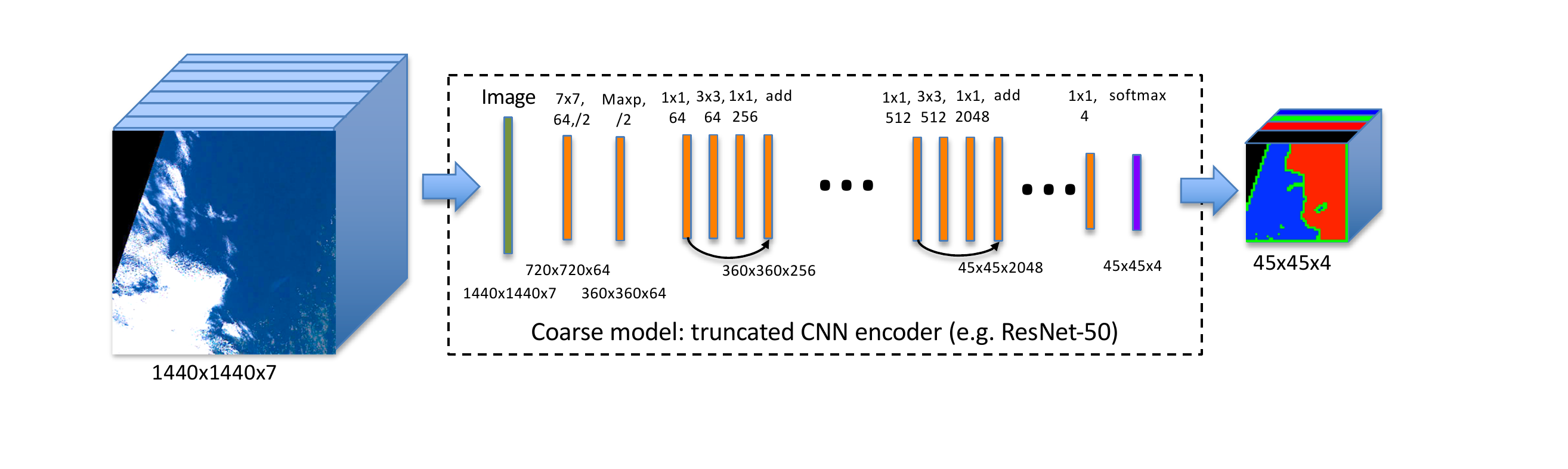} %
  \caption{The coarse component is built based on a CNN encoder, e.g. VGG, Inception or ResNet (illustrated) whose bottleneck layer is dimensioned to a size of e.g. 45x45, segmenting MSI images undersampled to 1440x1440 resolution, into a 45x45 grid of patches each assigned one of four classes.}
  \label{coarse_arch}
\end{figure*}

\subsection{Problem formulation}

Given a collection of $N$ high-resolution multispectral images represented as tensors $X_{n} \in \mathbb{N}^{h \times w \times b}$ and a ground truth\footnote{We use \textit{ground truth} to refer to the target annotation, following machine learning nomenclature. This is not to be confused with in-situ ground-level observations sometimes used as targets; we only use  remote sensing data.}
binary segmentation $y \in \{0,1\}^{h \times w}$ provided at pixel level for each of the images, the goal is to learn a neural network that can segment future images in a manner that accurately captures the properties of a ground truth segmentation with particular characteristics: In addition to interpreting the spectral composition of a small neighborhood of each pixel to denote clouds, the ground truth annotates a cloud-covered area with a preference for contiguous masks. In this work, we outline a scenario of learning binary cloud masks. However, the technical elements directly generalize to multi-class problems with a moderate number of classes, and are applicable to other domains of large images and contiguous segments.

Within this general description of supervised semantic segmentation problems, we focus on
\begin{itemize}
    \item Making improved use of both global and local information for increased accuracy.
    \item Learning from a ground truth that is not accurate at the level of individual pixel. Instead, the data is annotated with coarse contiguous areas of the occlusion class, so that small areas of background within the area are classified as occlusion.
    \item Good coverage (recall) of the occlusion class; some background classified as occlusion (by clouds) is acceptable, but not vice versa.
\end{itemize}

In the following, we first explain the overall model architecture in Section \ref{sec:architecture} and then provide the technical details for a loss function required for addressing the requirements of contiguity and emphasized recall in Section \ref{lossrationale}.

\subsection{Model architecture}
\label{sec:architecture}

Deep learning models are mostly trained using hardware acceleration units e.g. GPUs whose available memory per unit remains a practical bottleneck limitation \cite{chen2016training}\cite{wang2018superneurons} for processing large images despite constant 
improvements in processing speed. Common practice for processing conventional photograph-sized RGB images is to input undersampled images and adapt task objectives to the resulting low resolution of the outputs. For example, full-resolution segmentation is not necessarily required for applications such as identification and tracking of objects.
For high-resolution MSI images, however, the relative reduction of segmentation resolution becomes much greater, and significant loss of output resolution is undesirable, if not unacceptable, for many remote sensing applications.

Two main workarounds for processing high-resolution images are to (a) analyze undersampled images \cite{dronner2018fast} as described above, or (b) to analyze isolated smaller patches of the image, looping over multiple patches to process the whole image \cite{mateo2017convolutional}. Undersampling loses resolution and hence local information, but retains global information better and enables modeling of phenomena of larger spatial extent. Patching, in turn, parallelizes well and retains full resolution, but loses global information and hence has limited ability to model large spatial features. 
This is both because patching makes it impossible for a model to account for information outside the patch, but also because the geographical extent of CNN features is largely determined by the filter dimensions of the first layer, which is necessarily small when operating at the level of individual pixels.

To alleviate the drawbacks of either approach, we propose a model architecture, illustrated in Fig. \ref{graphstract}, that combines coarse analysis of undersampled images with fine-grained analysis for a small number of patches selected by the coarse model. This allows fast and memory-efficient analysis of global features, while retaining a capability for full-resolution segmentation.

We divide the MSI image logically into an e.g. $45\times45$ grid of constant-sized patches. This split is chosen to yield a manageable patch size for GPU training at full resolution for Sentinel-2 images, but can be easily adjusted to other image dimensions and available accelerator memory. The task of the \textit{coarse} component in our architecture is to provide a classification for each patch, exactly one out of four classes: "Overcast", "Partly Cloudy", "Cloudless", or "No Data" ("No Data" denotes missing data resulting from geospatial transformations from satellite imagery to the orthorectified tiles of Sentinel-2). The coarse model ingests an undersampled input of the complete image, and dimensions are selected so that each patch of the original image corresponds to a cell in the output grid of the coarse model. 

Of all patches, only those that were assigned to "Partly Cloudy" are segmented at full resolution with the \textit{fine-grained} component to pixel-level classes of "Cloud", "Clear", and "No Data". 
To avoid confusion, we use distinct class names at patch and pixel level (excluding "No Data"). The detailed cloud status, denoted by "Cloud" and "Clear", is available only at the pixel level. At the patch level, "Overcast" refers to a patch for which all pixels are classified as "Cloud". "Cloudless" refers to a patch for which all pixels are "Clear". Finally, "Partly Cloudy" corresponds to a patch having both "Cloud" and "Clear" pixels, i.e. needing more detailed segmentation.

\subsubsection{Coarse model}
\label{ch_coarse_model}
The coarse model component (Fig.~\ref{coarse_arch}) processes undersampled but spatially complete images. For this, we use the layers of an interchangeable CNN encoder (several are evaluated later), down to the narrowest layer that retains a 2d spatial shape in an image classifier or an autoencoder, i.e. the "bottleneck" layer. We replace the rest of the layers with a dimensionality-reducing $1 \times 1$ convolution and a softmax layer to obtain a single classification for each patch of our image on a $45\times45$ grid of patches.

A core property of the coarse model is its ability to account for features having a large spatial geographical extent. It analyzes images undersampled from 10980 to 1440 in both dimensions (see Section~\ref{data} and the Supplementary material for details), which means that any convolutional filter covers a roughly 60 times larger spatial area than the corresponding filter would if directly applied at the original full resolution of Sentinel-2 images.

\subsubsection{Fine-grained model}

The fine-grained model component can assume any given pixel-classifying semantic segmentation CNN architecture that is able to operate on $256\times256$ patches sliced at full resolution from the input image (see Supplement for a representative detailed example on a UNet with a ResNet-50 backbone). The fine-grained model is distinct and separate from the coarse model. In Section~\ref{ch_results}, we present comparative results for several alternatives, eventually selecting FPN architecture with a SEResNeXt-50 encoder backbone.  Patch size includes an overlap of 4 pixels with the adjacent patch to minimize patch edge artifacts. Thus, predictions are cropped to a patch size of $248\times248$. We use a binary cross-entropy loss and a post-processing threshold. Although binary cross entropy is designed to allow multi-class labels per pixel, this choice empirically outperformed categorical cross entropy in our experiments.

\subsection{Coarse model loss function for emphasizing a boundary}
\label{lossrationale}

\begin{figure}[t!]
  \begin{tabular}{ccc}
  \includegraphics[width=0.26\columnwidth]{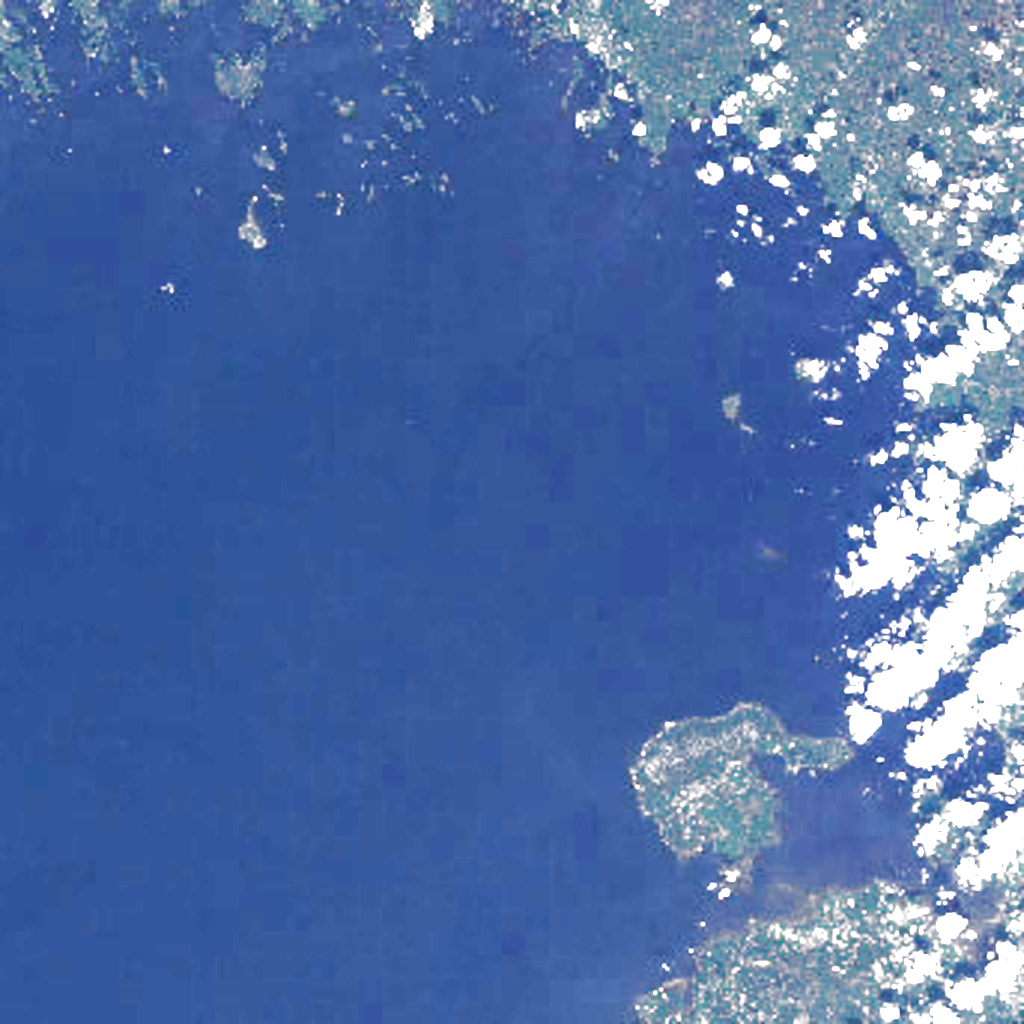} &
  \includegraphics[width=0.26\columnwidth]{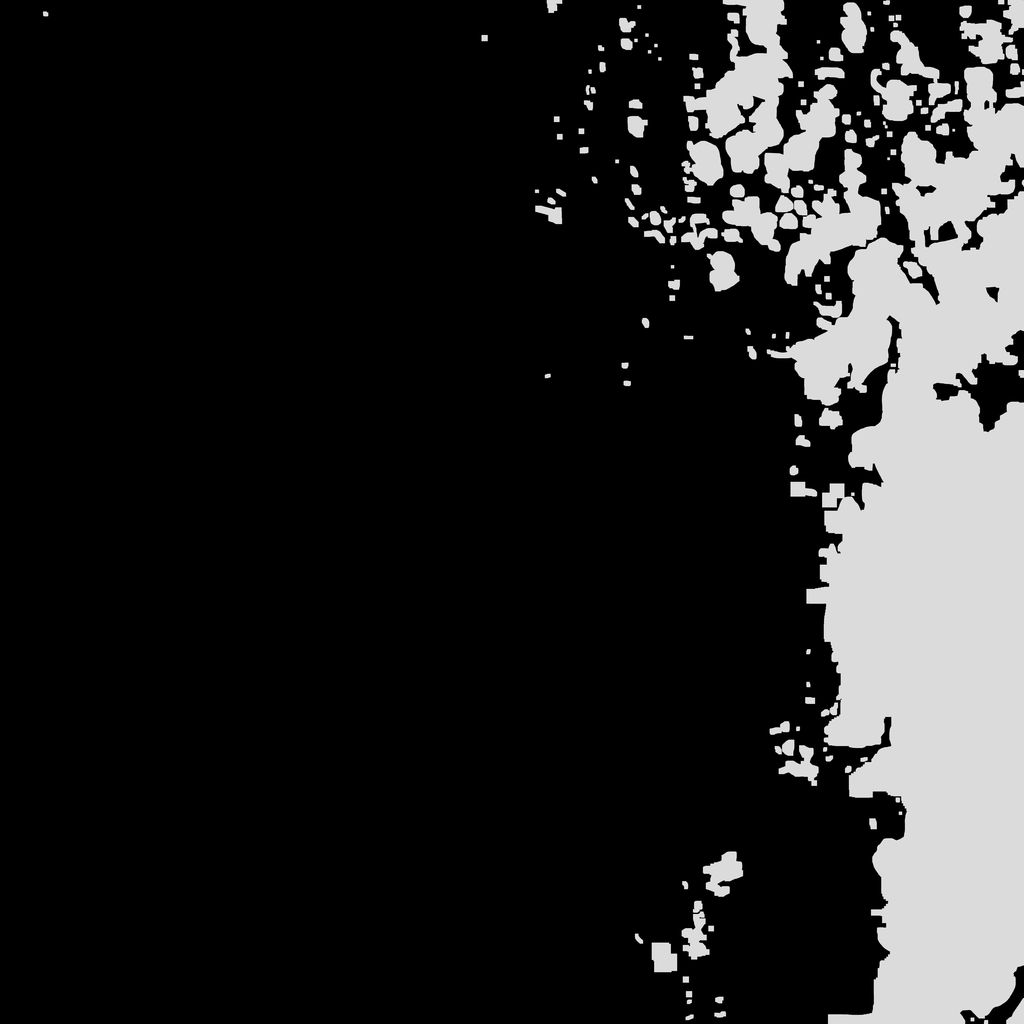} &
  \includegraphics[width=0.26\columnwidth]{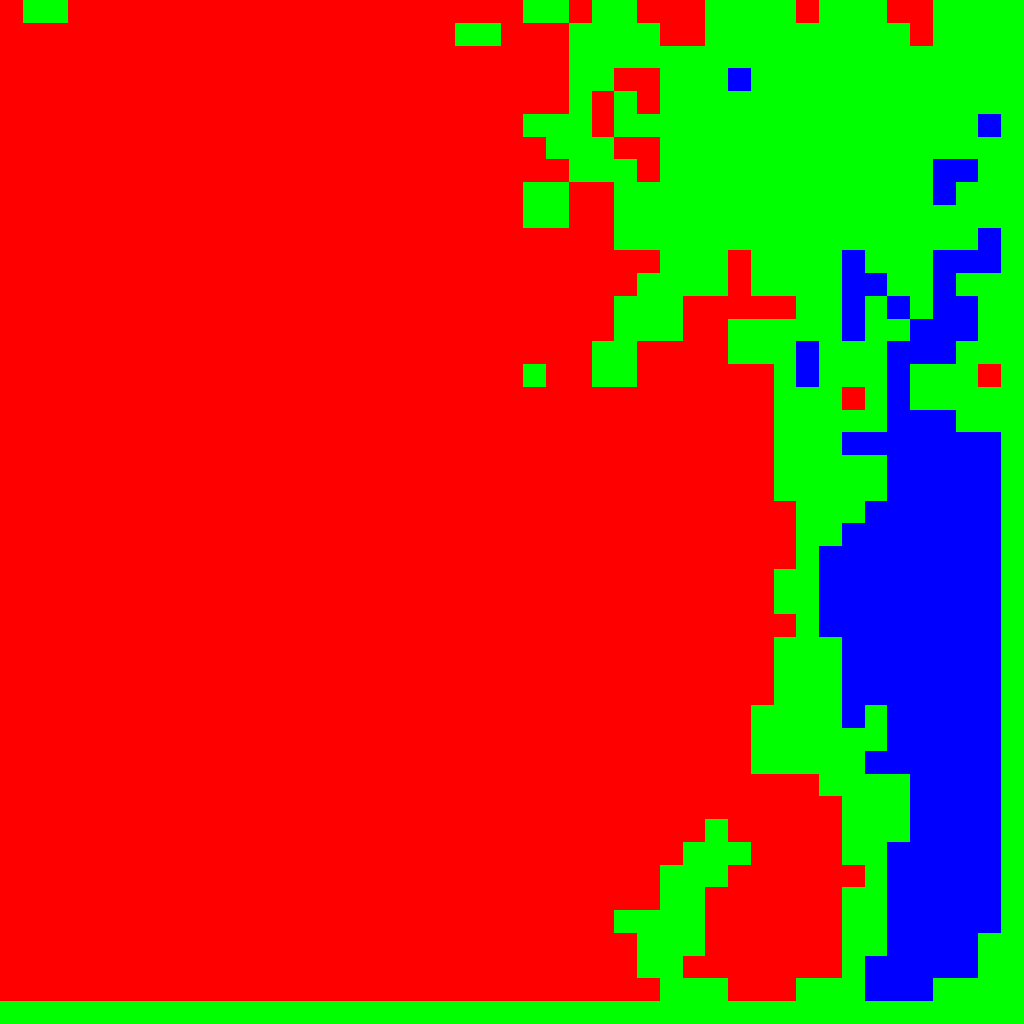} \\
  \includegraphics[width=0.26\columnwidth]{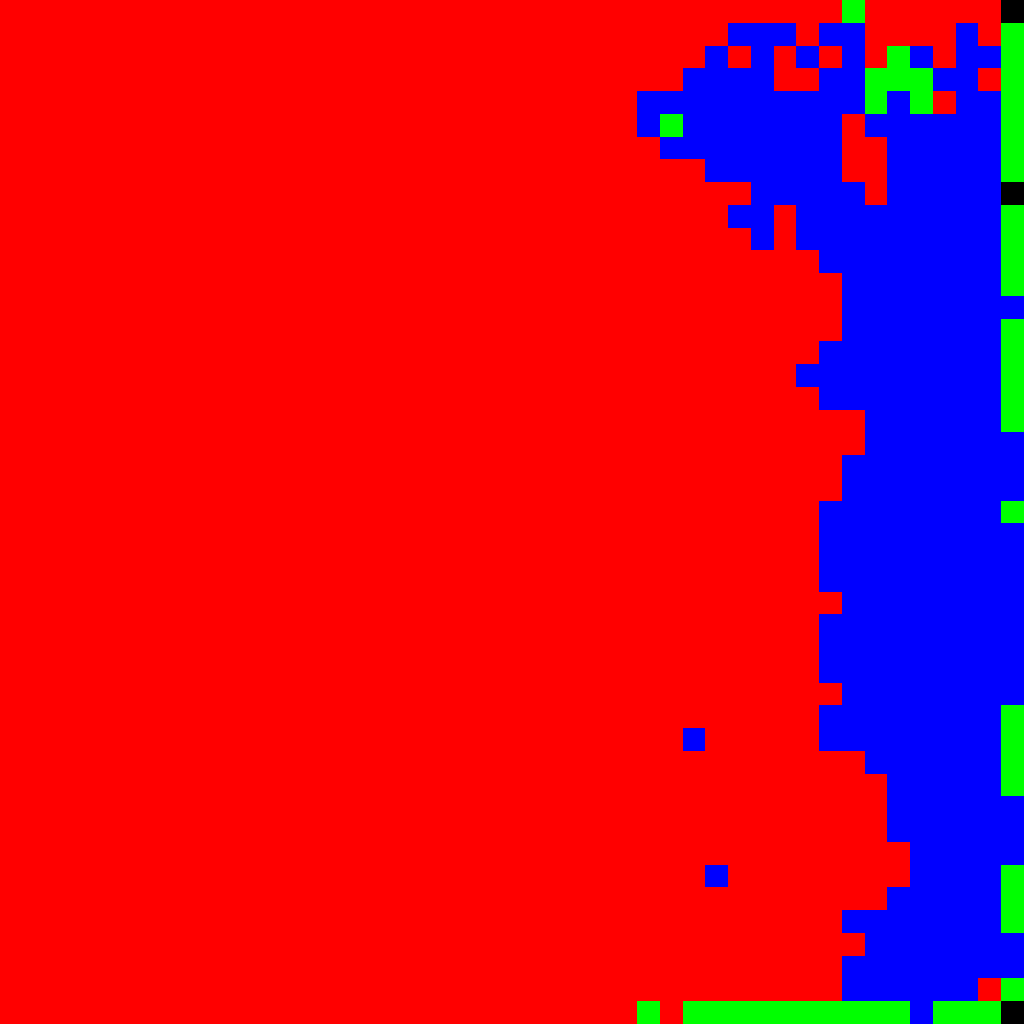} &
  \includegraphics[width=0.26\columnwidth]{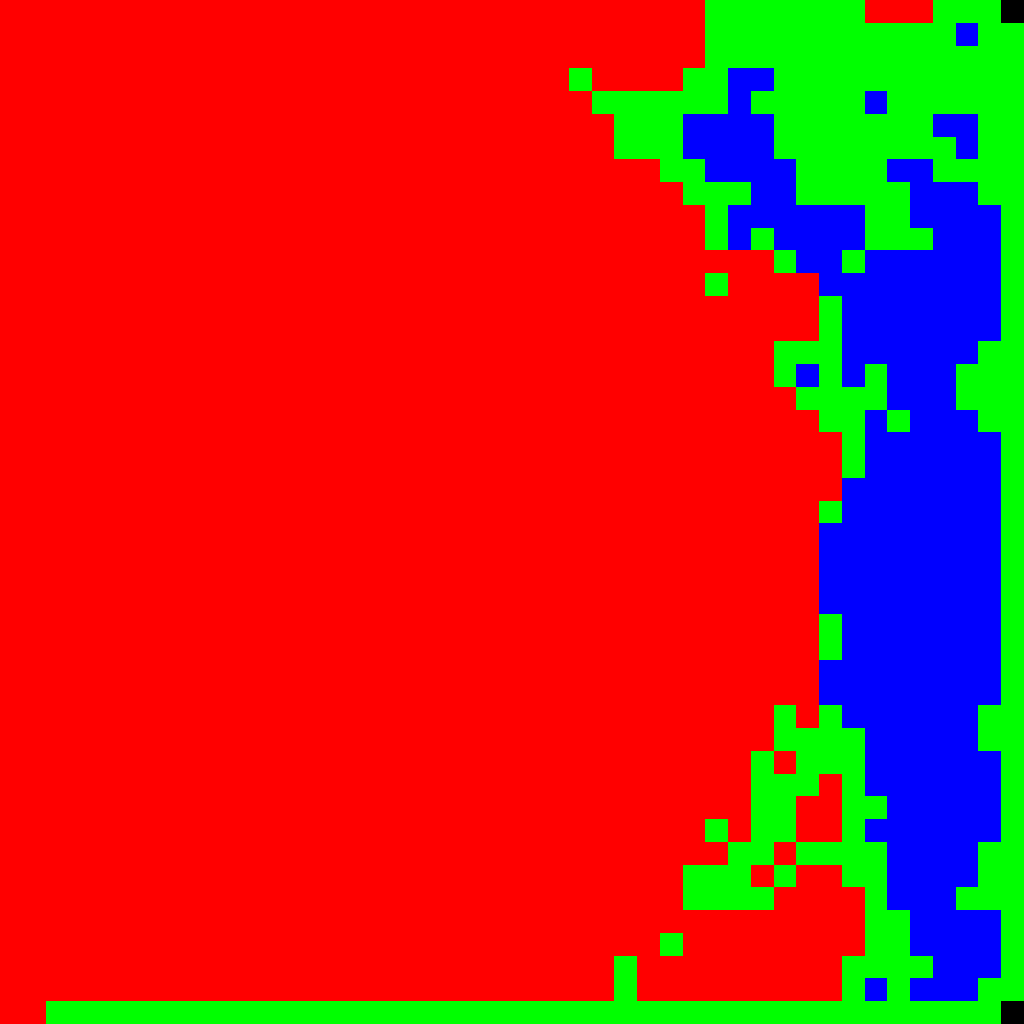} &
  \includegraphics[width=0.26\columnwidth]{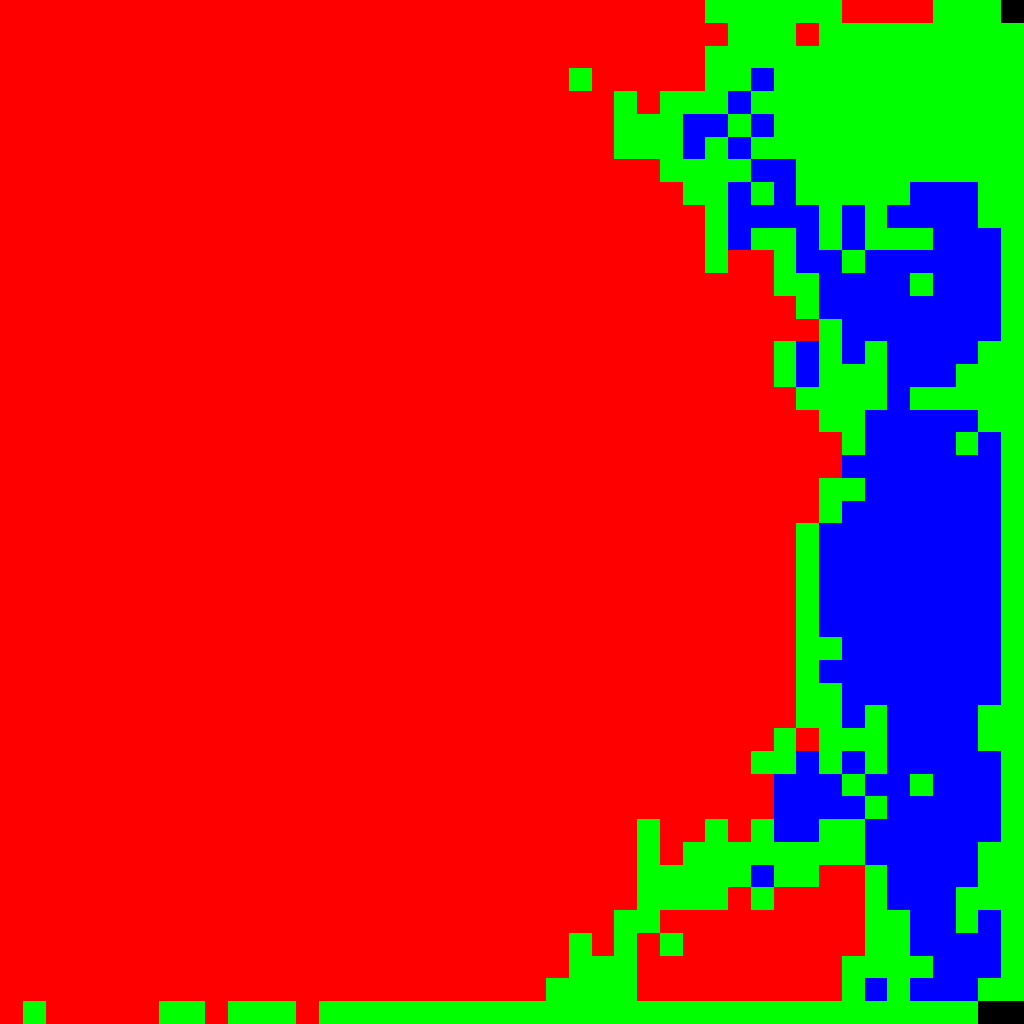}
  \end{tabular}
  \caption{
  Coarse model loss function rationale. \textbf{Top row} shows the original Sentinel-2 MSI image (left), the binary ground truth mask (middle), and the derived 45x45 ground truth for training the coarse model (right). Here 
  "Partly Cloudy" (green) patches have both "Cloud" and "Clear" pixels, whereas for the other two classes ("Overcast" for blue, "Cloudless" for red) all pixels belong to the same class. 
  \textbf{Bottom row} illustrates the predictions of the coarse model for three loss functions. 
  A model trained for unweighted categorical cross entropy (left) completely fails to reproduce the boundary. Using weighted cross entropy helps in recovering the border (middle), and encouraging predictions with correct number of border patches with the adjacency loss (right) captures the border best. 
  }
    \label{fig_lossrationale}
\end{figure}

For the coarse model component, we want to bring out the "Partly Cloudy" class visible at the cloud boundaries of the derived ground truth (Fig.\ref{fig_lossrationale}), whereas the uniformly masked patches of "Overcast" and "Cloudless" as such provide contiguity to the inner and outer regions of cloud segments.  To achieve this, we propose a loss function that can adjust for recall or precision of a class as well as measure and replicate a class' adjacency with other classes against the ground truth.

For all but the very smallest shapes, a raster perimeter drawn around the shape mostly covers a smaller area than the inner or outer area of the shape. This explains the uneven class distribution: the border class ("Partly Cloudy") between fully cloudy ("Overcast") and ("Cloudless") areas is small. The boundary consists of patches that contain both cloud and clear pixels. To improve detection of the boundary, we (a) set weights on weighted cross entropy of each class, and (b) encourage segmentations for which the count of the border patches closely matches the count in ground truth. Intuitively, we would rather allow a small extra amount of  pixel-level processing along the borders in between than totally miss a border region between "Overcast" and "Cloudless".

To detect the boundary patches, we optimize for a two-term loss function consisting of 
a custom \textit{adjacency loss} term, $\mathcal{L}_{\mathrm{adj}}(y, \hat{y}) $, and a \textit{weighted cross-entropy} term, $\mathcal{L}_{\mathrm{wce}}(y, \hat{y})$. The overall loss to be minimized is
\begin{gather}
\mathcal{L}(y, \hat{y}) = \gamma \mathcal{L}_{\mathrm{adj}}(y, \hat{y}) + \mathcal{L}_{\mathrm{wce}}(y, \hat{y}), 
\label{eq1}
\end{gather}
where $\gamma \in \mathbbm{R}^{+}$ is an adjustable weight between the two terms, y is the ground truth binary indicator tensor and $\hat{y}$ is a predicted probability tensor, both y:s 
indicating class membership.
\begin{align}
\mathcal{L}_{\mathrm{wce}}(y, \hat{y}) = -  \sum_{k=1}^{C}\alpha_{k}\sum_{i=1, j=1}^{m, n}
\label{weightedcrossentropy}
&[{y}_{ijk} \mathrm{log}(\hat{y}_{ijk})~+ \\
&\beta_{k}(1-y_{ijk})~\mathrm{log}(1-\hat{y}_{ijk})]~,\nonumber
\end{align}
The second term, weighted cross entropy (Eq. \ref{weightedcrossentropy}), controls the ratio of each class with a class-specific weights $\alpha_k \in [0,1]$ and $\beta_{k} \in [0, \infty[$. 
C=4 is the number of coarse model classes, $\alpha_{k} \in [0, 1]$ the importance weight s.t. $\sum_{k=1}^{C}\alpha_{k} = 1$, and $\beta_{k} \in [0, \infty[$ is the cross-entropy weight that should be $<1$ to reward recall of class k. 
The values for the hyperparameters used in the experiments are given in Section \ref{ch_results}.

The first term, \textit{adjacency loss}, is designed so that it is minimized when the number of adjacencies between any two classes matches between the prediction and the ground truth, in order to emphasize solutions that accurately model class borders. We denote by $S'(y)$ an \textit{adjacency score} that counts the number of adjacencies and define the \textit{adjacency loss} as the squared difference between the prediction and the ground truth
\begin{gather}
\mathcal{L}_{\mathrm{adj}}(y, \hat{y}) = [S'_{\mathrm{d}}(y) - S'_{\mathrm{d}}(\hat{y}]^2.
\end{gather}
Regularization of segmentation results based on adjacencies has long history in image processing, typically in form of directly penalizing for adjacency of certain classes, either by a Markov random field or specific loss terms such as the one recently proposed by Ganaye et al. \cite{ganaye2018semi}.
Our formulation is conceptually very different: We do not penalize for adjacency of any classes as such, but instead penalize for a difference in the count of adjacencies between the prediction and the ground truth. This allows finer control of the segmentation result.

\begin{figure}[t!]
  \includegraphics[width=1.0\columnwidth, trim=0 9cm 9.5cm 0, clip, center]{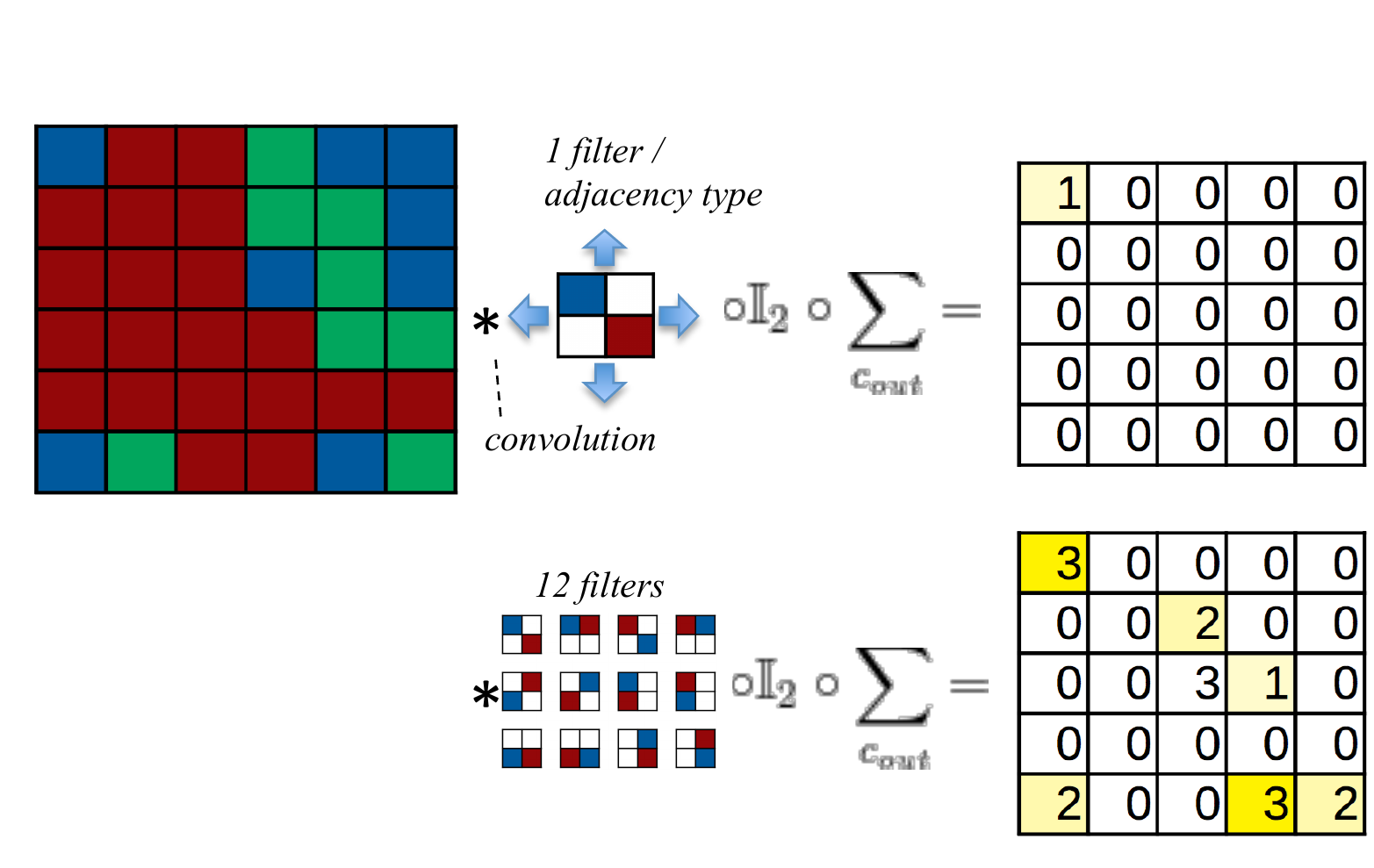} 
  \caption{Adjacency score counts instances of two adjacent pixels on two bands. The score is calculated by moving twelve 2x2 convolution filters for $c_{\mathrm{in}}=2$ bands over the image, one filter for each adjacency type. Each filter's convolution at two adjacent pixels produces a value of 2 (not shown in the picture, only the count of 2's), hence the indicator function. The highlighted numbers indicate the total count of adjacencies detected by the indicator for all $c_{\mathrm{out}}=12$ filters.}
  \label{fig_combinations}
\end{figure}

Even though we will eventually need a differentiable loss for optimization purposes, we start by defining a \textit{discrete adjacency score} $S_{\mathrm{d}}(y)$ (Eq.\ref{eq_adjscore}) that counts the total number of adjacently located instances of two pixel classes. For any two classes ($c_{\mathrm{in}}=2$), the score can be computed using $c_{\mathrm{out}}=12$  convolutional $2\times2$ kernel filters $K_k$, corresponding to the 12 possible adjacency relations (see Figure~\ref{fig_combinations}), using
\begin{align}
S_{\mathrm{d}}(\hat{y}) = & \sum_{i=1, j=1, k=1}^{m-1, n-1,c_{\mathrm{out}}}\mathbbm{1}[\mathrm{\sigma}(\hat{y})\ast K_k = 2] 
\label{eq_adjscore}
\end{align}
Here $\hat{y}$ is a probability tensor, e.g. a two-class subset of a 2D multi-class membership probability mask as output by e.g. a $softmax$ activation in a CNN, of dimension $m\times n\times (c_{\mathrm{in}}=2)$, $\mathrm{\sigma}$ is a one-hot function turning a real-valued set of vectors into one-hot binary format, and * is the n-dimensional convolution operator here assuming a stride of 1 and no padding. 
$y_{oh}= \mathrm{\sigma}(\hat{y})$ then corresponds to a set of (here, two) mutually exclusive binary 2D pixel masks stacked depth-wise, i.e. a one-hot format of a semantic segmentation, that can be a ground truth or a prediction, of dimension $m\times n\times c_{\mathrm{in}}$ , where $m$ and $n$ are width of and height in pixels of a segmentation prediction of an image.
Finally, an indicator function $\mathbbm{1}$ followed by a summation over $c_{\mathrm{out}}=12$ filters together count the number of positions that have a value corresponding to a detected adjacency on $(c_{\mathrm{in}} =) 2$ bands, i.e. a value of 2.
For a normalized, more generalizable metric, we divide $S_{\mathrm{d}}(\hat{y})$ by the amount of pixels and the maximum number $A_{\mathrm{max}}$ of simultaneous adjacency types per filter area (Eq.\ref{normscore}). For example, when filter dimensions $m=n=2$ for a single offending pixel with no appropriate boundary region, $A_{\mathrm{max}}=4$ (Fig.\ref{fig_combinations}).
\begin{equation}
S_{\mathrm{dn}}(\hat{y}) =  \dfrac{S_{\mathrm{d}}(\hat{y})}{mnA_{\mathrm{max}}}.
\label{normscore}
\end{equation}

Finally, we convert the discrete adjacency score into a differentiable loss for training the model, by approximating the indicator function $\mathbbm{1}$ with the rectified linear unit (ReLU) and a threshold $1+\epsilon$: 
\begin{align}
\label{continuousadjloss}
S'_{\mathrm{d}}(\hat{y}) = &\sum_{i, j, k}^{m-1, n-1,c_{\mathrm{out}}}\mathrm{ReLU}(\mathrm{\sigma}'(\hat{y})\ast K_k - (1+\epsilon)) \\
s.t. \quad & y'_{oh} = \mathrm{\sigma}'(\hat{y}) \in \{\{0+\rho_{1}, 1 - \rho_{2}\}^{c_{\mathrm{in}}}\}^{m \times 
n}  , \nonumber\\ 
&\sum_{c=1}^{c_{\mathrm{in}}}[\mathrm{\sigma}'(\hat{y})]_{c} \approx 1.\nonumber
\end{align}
Here $\epsilon$ and $\rho$ are small residuals, and $\mathrm{\sigma}'()$ is a continuous-valued relaxation of one-hot encoding (see Supplement for details). We set $\epsilon$ to $0.001$ and $\rho$ at floating point precision.

\section{Data}
\label{data}
\begin{figure}[t!]
  \includegraphics[width=0.45\textwidth, center]{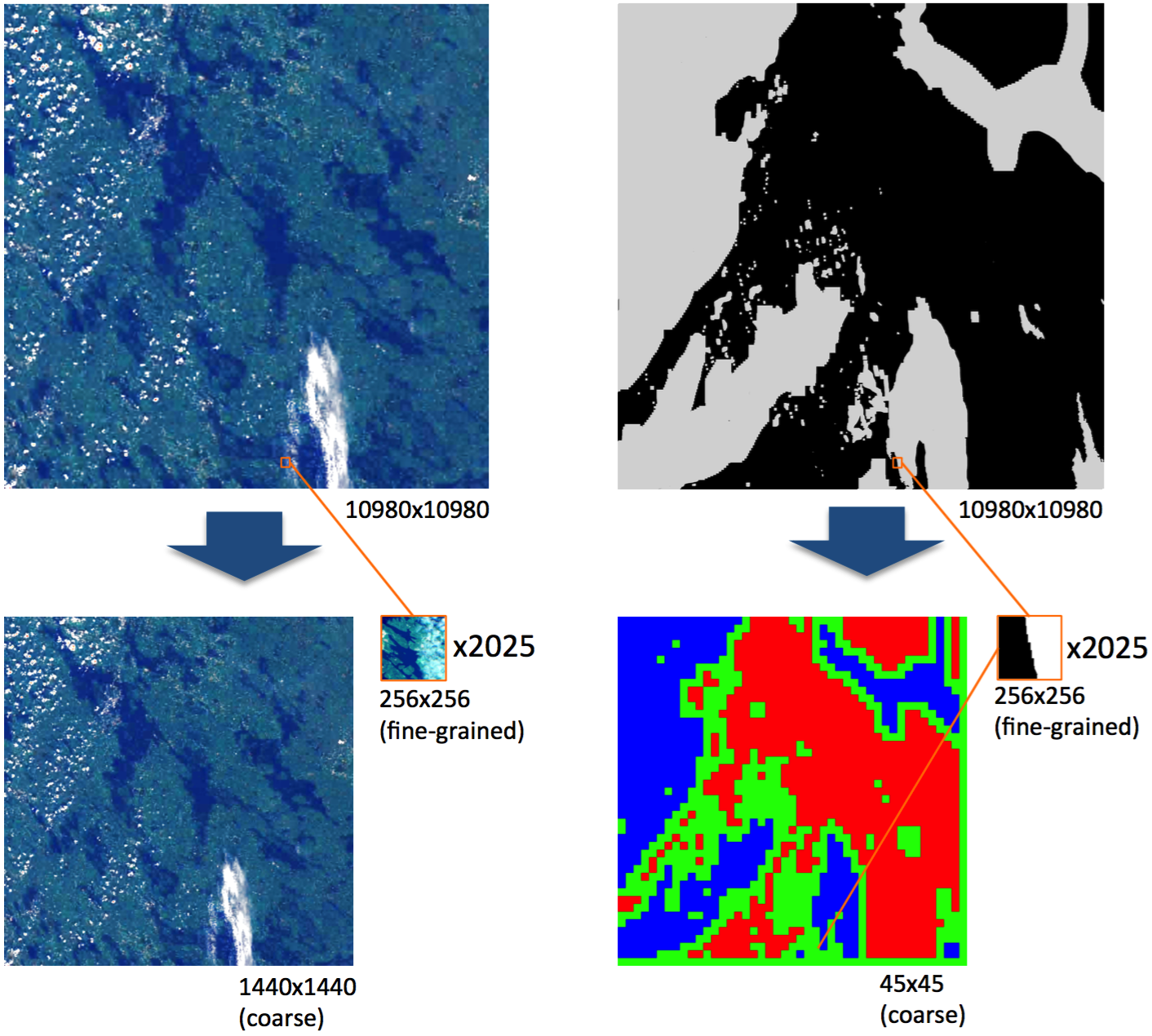}
  \caption{Illustration of the data pipeline for a single MSI image. \textbf{Top left:} Original MSI image data (only visible RGB bands shown here). \textbf{Top right:} fine-grained model full-resolution annotations for classes "masked"/ "accepted", \textbf{Bottom left:}  coarse model data downsampled to 1440x1440 and fine-grained model data as overlapping 256x256 patches. \textbf{Bottom right:} for the coarse model, annotations are pooled to a 45x45 grid from full resolution, resulting in classes "Overcast" (blue), "Partly Cloudy" (green) and "Cloudless" (red). For the fine-grained model, overlapping 256x256 patches are extracted from annotations ("No Data" regions omitted from illustration).}
  \label{groundtruths_fig}
\end{figure}

For training and evaluating the model,
we use a dataset of 478 Sentinel-2 (S2) MSI L1C images from years 2016 to 2017, manually selected for a minimal cloud cover and annotated for remaining cloud masks at the Finnish Environment Institute (SYKE). These same masked images have been combined to construct a cloudless mosaic and used as raw data for producing the EU Corine land cover mapping of Finland \cite{corinelc}. 

\subsection{Annotation}

Pixel-level annotation work took several months of work by a team of three annotators, using an initial reference mask produced by the Idepix tool of the SNAP software package by ESA as a basis (masks used were \textit{f\_cirrus\_sure, f\_cirrus\_ambiguous, f\_brightwhite}). As MSI L1C bands 1-4 and 8-10 were known to contain the most relevant information for cloud masking, they were viewed simultaneously in different false-color combinations to serve as the annotators' additional reference input.  As the annotation routine and composition of the team evolved and changed during the course of the annotation work, variability was introduced in the level of detail and features of the resulting manual masks. Spring and winter images contain snow and ice annotated to the same class as clouds since the pixels were unusable from the viewpoint of annotators, considering the targeted land-cover map (there is no permanent snow cover in Finland). Also from the data and algorithm viewpoint, clouds can be a challenge to discern from snow and ice \cite{cloudmask_assess}.

The annotators were instructed to mask any and all cloud-contaminated areas including partial and sparse occlusion as "Cloud", which in practice results in cloud overestimation. Examining and annotating every pixel in detail would have been an overwhelming and practically infeasible task, even with these relatively ample annotation resources given the size of the dataset, $478 \cdot 10980^2 \approx 58\cdot10^9$ multispectral pixels. 
Since snow was treated as an unusable class just like clouds for the land use application in question, snow-covered areas were likewise annotated as "Cloud".  Finally, the cloud edges were automatically extended with a buffer of approximately 10 pixels wide to include mixed pixels in the resulting mask (see Figure 6). Also, "No Data" areas of MSI images were included as part of the "Cloud" class. 

\subsection{Data pipeline}
\label{data_pipeline}

In the following, we briefly outline the data processing pipeline used for transforming the original MSI data and the ground truth annotations to form the training and evaluation data sets for the model. Full details of the pipeline are provided in the Supplement.

Using the original two binary classes of the annotations, i.e. "Clear" and "Cloud",
we construct three mutually exclusive full-resolution ground truth pixel classes: "Cloud", "Clear" and a reconstructed class for "No Data". "No Data" corresponds to invalid polygonal image regions not covered by the satellite sensors.
In the end, we have 10980x10980x7 spectral data with 16-bit depth and a 10980x10980x3 boolean ground truth mask.

Using the outcome of this initial processing step, we derive two separate datasets, one for the coarse and the fine-grained model each. 
For the coarse model, each MSI image is downsampled to 1440x1440x7, and ground truth masks of 45x45x4 with four classes are derived for patches of the original data so that patches with only "Cloud" pixels are annotated as "Overcast" and patches with only "Clear" pixels as "Cloudless", whereas patches with both are marked as "Partly Cloudy".
We augment the coarse model dataset by simple vertical and horizontal flips to increase the data volume, resulting in a total of 1912 images associated with ground truth masks.

For the fine-grained model, we slice each MSI image to dimensions of 45x45x256x256x7. Altogether 478 images yield a total of 975,950 patches of spectral dimensions of 256x256x7 with an overlap of four pixels. Correspondingly, a ground truth mask of 45x45x256x256x3 mask is extracted for the fine-grained model. 
Due to the contiguous nature of the ground truth, the share of fully "Overcast" or "Cloudless" patches is disproportionately high as noted in \ref{lossrationale}. For training the fine-grained model, we downsample the amount of overcast and cloudless, having first divided patches into ten bins by their respective ratio of ground truth "Cloud" pixels. In a natural distribution without weighting, the bins at the extremes of $<10\%$ and $>90\%$ of cloudiness ratio have a proportion of more than ten times that of the patches in other bins that are partially cloudy. Naively training on the whole data would bias the model towards outputting patches that have pixel classes of almost uniform "Cloud" or "Clear", yet in our framework, the fine-grained model is only used for patches already known and classified to be only partly cloudy. To remove this bias, we re-weight the ten bins so that the extreme bins have a weight of $16.7\%$ and the remaining eight all have a weight of $8.3\%$.

\subsection{Model training}

The models were trained on 454 of the original 478 MSI images. For the fine-grained model, a proportion of patches from the training images was separated and held out from training into a validation set for estimating optimal hyperparameters. The remaining randomly selected 24 whole images were held out for model testing evaluation, i.e. not used for training or development of the models. 

We implemented the proposed architecture with Keras and TensorFlow, using standard stochastic gradient descent with a momentum of 0.9 for training both model components.  We initially trained the coarse model from scratch for 131 epochs over 9 hours with weighted cross-entropy $\mathcal{L}_{\mathrm{wce}}$ only, i.e. with $\gamma=0$. This provides a baseline for $\mathcal{L}_{\mathrm{wce}}$ as such for loss function comparison in Section \ref{exp2}. For training with the adjacency loss, we use transfer learning, i.e. assume the weights of the $\mathcal{L}_{\mathrm{wce}}$ baseline as a starting point, activate the $\mathcal{L}_{\mathrm{adj}}$ term with $\gamma=1$ and train for additional 14 epochs to get the comparison metrics for $\mathcal{L}_{\mathrm{adj}}$.
The fine-grained FPN model took 40 epochs over 5 hours to converge from scratch on a random sample of 8000 patches per epoch (500 mini-batches x 16 image patches) out of approx. 1 million patches. We used a threshold of 0.45 for the "Cloud" pixel class, the threshold optimized against a separate validation set. 

\section{Results and discussion}
\label{ch_results}

\begin{figure*}[t!]
    \begin{minipage}[b]{.15\textwidth}
    \centering
      \begin{tabular}{l}
        \includegraphics[width=57pt, center]{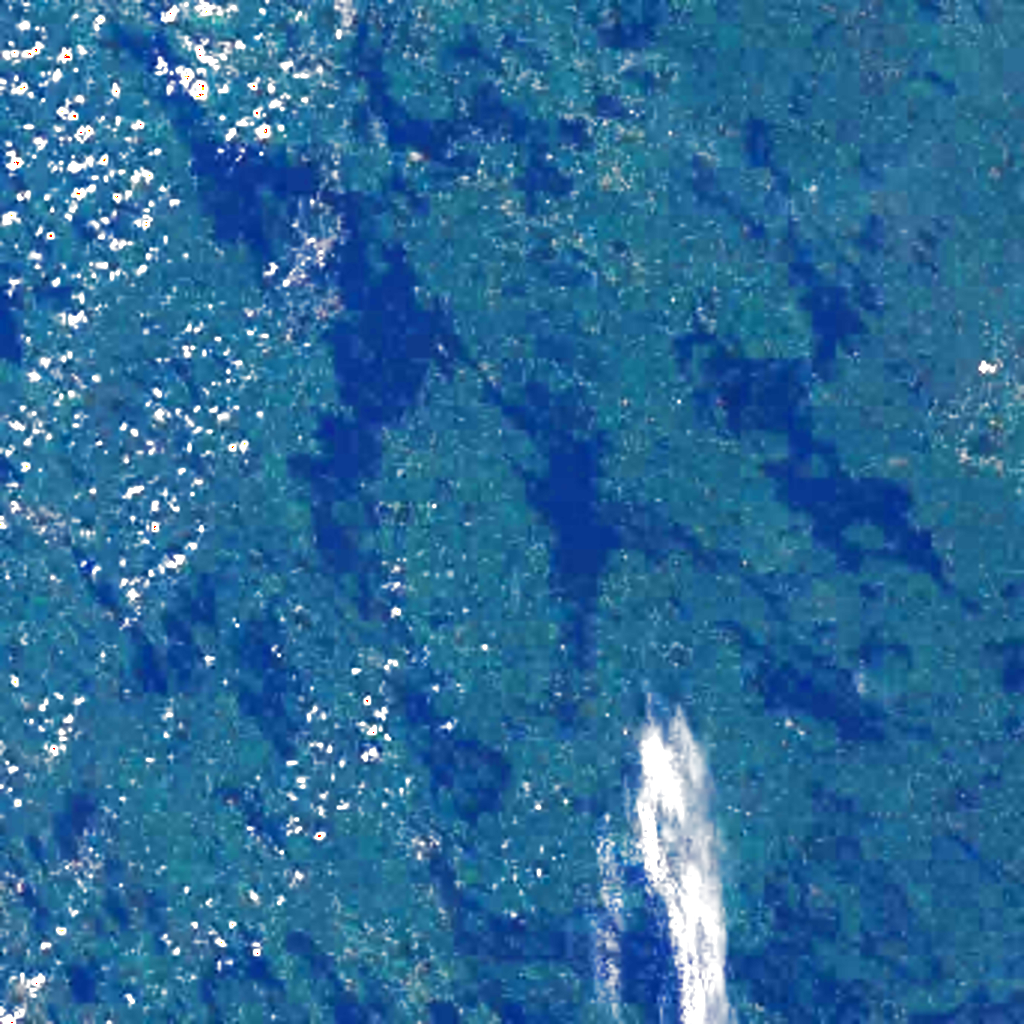} \\ \\ \includegraphics[width=57pt, center]{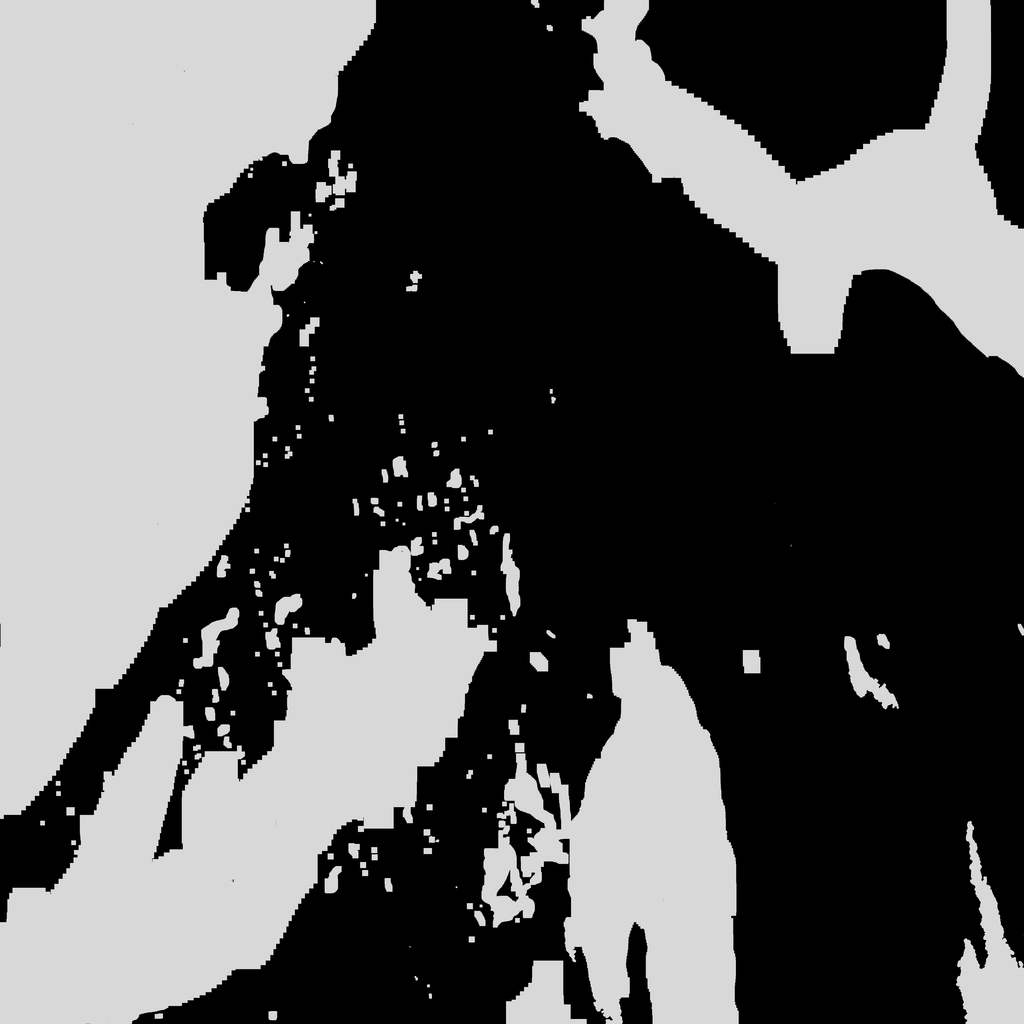}
      \end{tabular}
    \end{minipage}%
    \begin{minipage}[b]{.8\textwidth}
    \centering
      \begin{tabular}{ccc}
        \includegraphics[width=130pt]{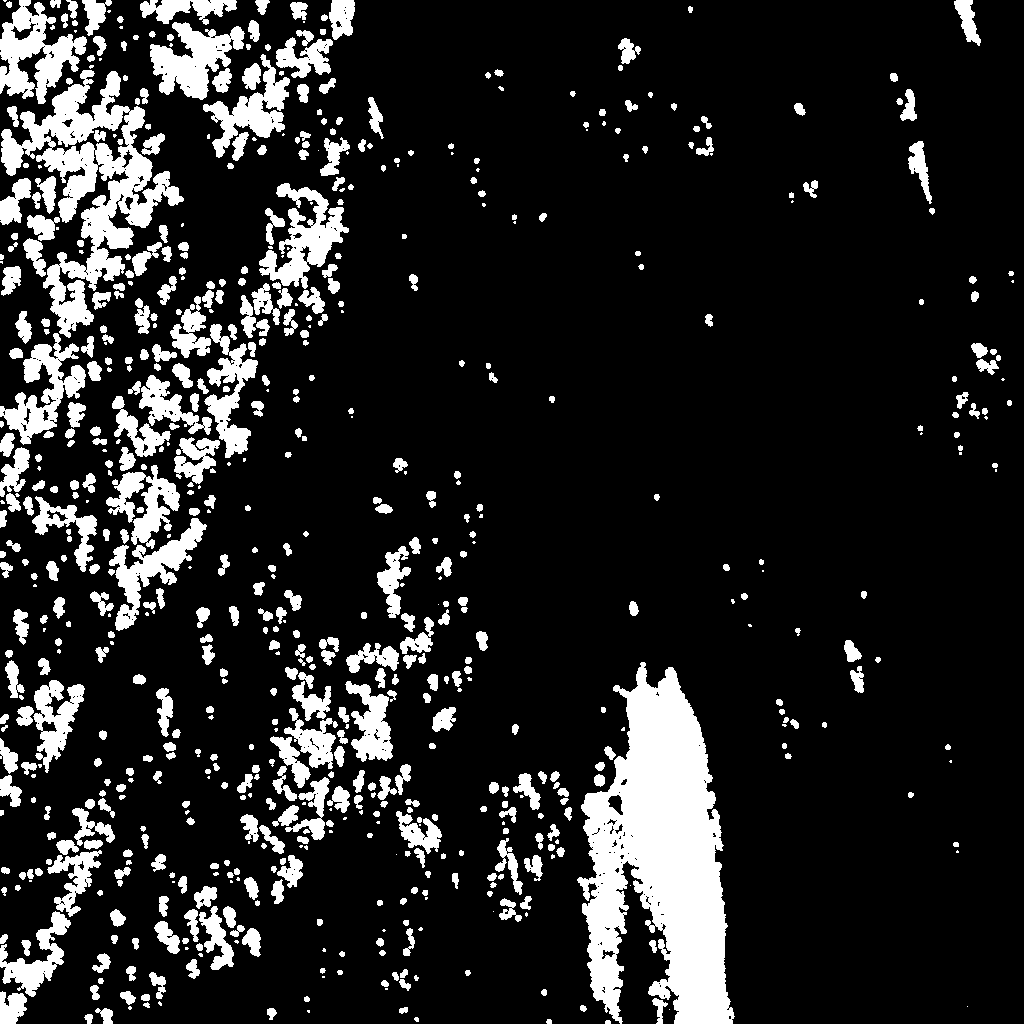} & \includegraphics[width=130pt]{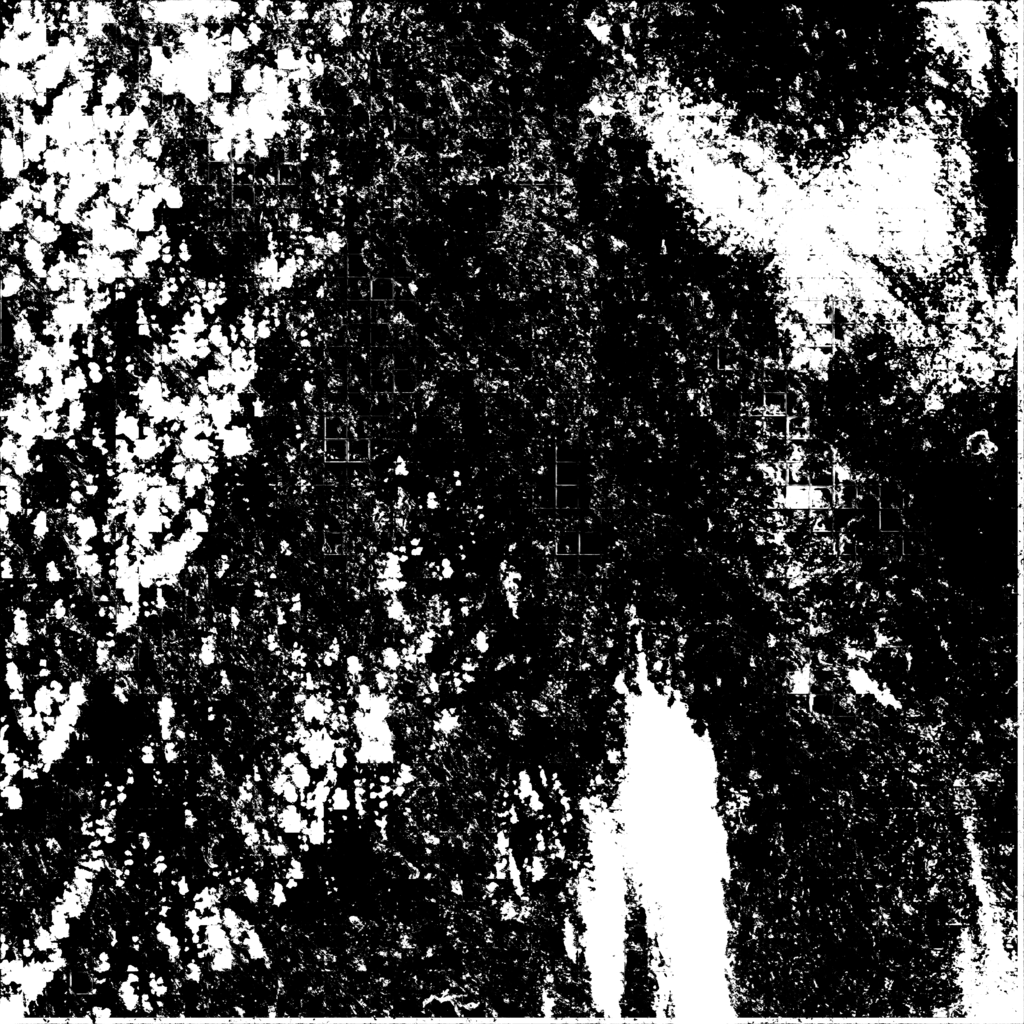} & \includegraphics[width=130pt]{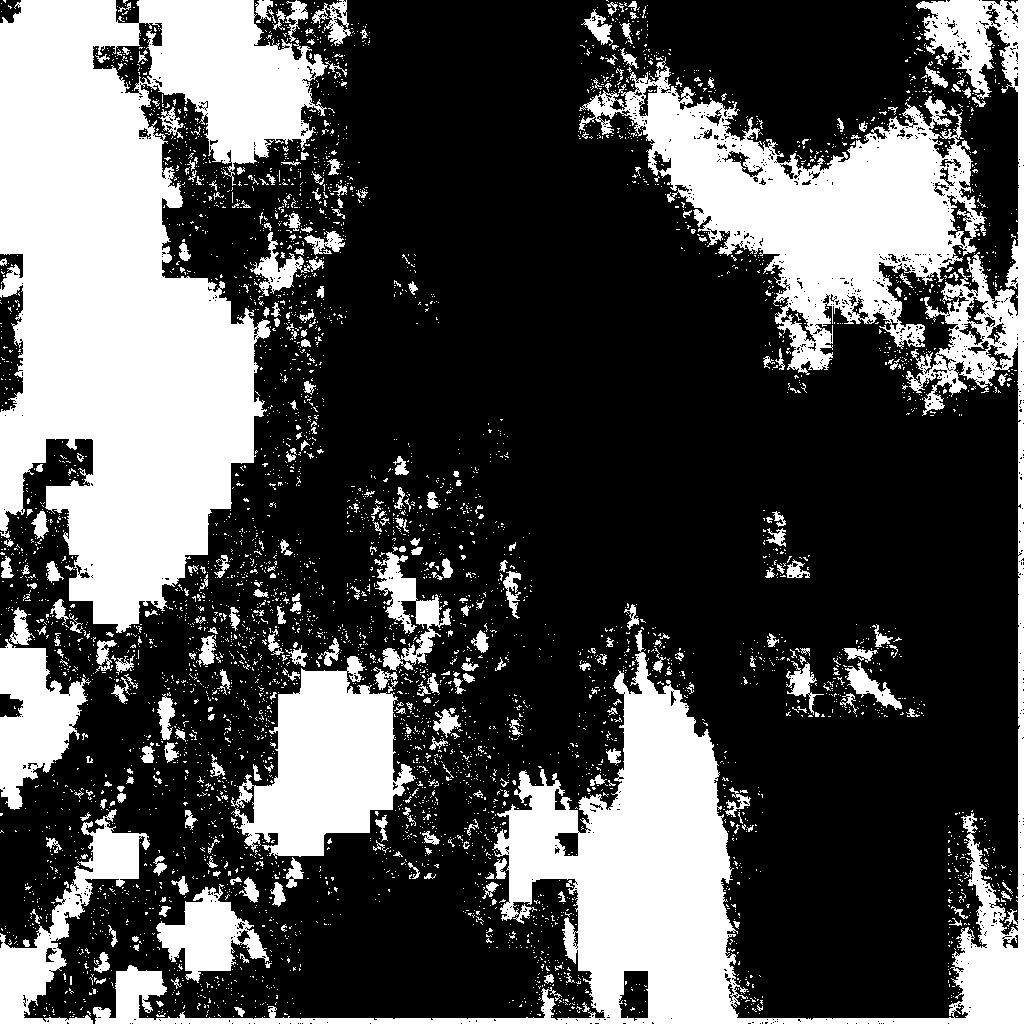}
      \end{tabular}
    \end{minipage}%
    \caption{Segmentation results. The small images to the extreme left show the MSI image and the binary ground truth, followed by the masks produced by the FMASK cloud detection model (left), best CNN baseline (center) and the proposed model (right). The proposed model clearly matches the ground truth most accurately of the three.}
    \label{fig_results}
\end{figure*}

We measure and evaluate the performance of our framework from the points of view of CNN-based semantic segmentation and MSI cloud detection algorithms. As our framework addresses the problem of an imbalanced dataset partly by a custom loss function, we additionally compare loss function alternatives to demonstrate the justification for our choice.

\subsection{Metrics}
\label{sect_metrics}

Metrics used for mutual comparison are accuracy ($\frac{TP+TN}{TP+TN+FP+FN}$), recall ($\frac{TP}{TP+FN}$), precision ($\frac{TP}{TP+FP}$), IoU ($\frac{TP}{TP+FP+FN}$), and F1 score ($\frac{2TP}{2TP+FP+FN}$), measured for full resolution pixels of the ground truth vs. model prediction. T for True and F for False tells if a pixel is classified correctly or incorrectly.  P stands for Positive i.e. "Cloud" pixel, while N stands for Negative i.e. "Clear" pixel. So, e.g. True Positive (TP) corresponds to the number of pixel instances that were correctly classified as "Cloud".  All metrics are mean values measured over a test set of randomly selected 24 images held out from the training set. 

\subsection{Baselines}
\label{sect_baselines}

\begin{table}[t!]
\caption{Comparison of encoder backbones used with a semantic segmentation architecture initially fixed to UNet. Boldface indicates best method for each metric.}
\label{exp1_table}
\begin{tabular}{|l|l|l|l|l|l|l|}
\hline
\textbf{Fine encoder}                & \textbf{accuracy} & \textbf{precision} & \textbf{recall} & \textbf{f1}     & \textbf{iou}    \\ \hline \hline
EfficientNet          & 0.5515            & 0.5112             & \textbf{0.8308} & 0.5769          & 0.4686          \\ \hline
ResNet-50             & 0.7158            & 0.7044             & 0.6422          & 0.6001          & 0.4777          \\ \hline
VGG16                 & 0.7299            & \textbf{0.7219}    & 0.6122          & 0.6026          & 0.4817          \\ \hline
InceptionV3           & 0.7336            & 0.6936             & 0.6907          & \textbf{0.6295} & \textbf{0.508}  \\ \hline
SEResNeXt-50          & \textbf{0.7345}   & 0.6898             & 0.6703          & 0.6207          & 0.5006          \\ \hline
\end{tabular}
\end{table}

\begin{table}[t!]
\caption{Comparison of encoder-decoder semantic segmentation architectures, with the encoder backbone fixed to SEResNeXt-50 (see Table \ref{exp1_table}).}
\label{exp1_1_table}
\begin{tabular}{|l|l|l|l|l|l|l|}
\hline
\textbf{Fine segm.}                & \textbf{accuracy} & \textbf{precision} & \textbf{recall} & \textbf{f1}     & \textbf{iou}    \\ \hline \hline
UNet          & 0.7345   & 0.6898             & 0.6703          & 0.6207          & 0.5006          \\ \hline
LinkNet       & 0.6346            & 0.5562             & \textbf{0.736}  & 0.5807          & 0.4636          \\ \hline
PSPNet        & 0.6666            & 0.5841             & 0.7069          & 0.5861          & 0.4675          \\ \hline
FPN           & \textbf{0.737}    & \textbf{0.7184}    & 0.6426          & \textbf{0.6167} & \textbf{0.4935} \\ \hline
\end{tabular}
\end{table}

\begin{table}[t!]
\caption{Comparison of variants of the proposed multi-scale network with different coarse model encoders, both with fine-grained model set to the best CNN baseline of Table~\ref{exp1_1_table} (first three rows; with "Fine" assigned to SEResNeXt-50/FPN as the fine-grained network) and with naively mapping all "Partly Cloudy" patches as "Cloud" (next three rows) to standard cloud detection models tested against our dataset. For reference, we also report the best CNN baseline again.}
\label{exp1_2_table}
\begin{tabular}{|l|l|l|l|l|l|l|}
\hline
\textbf{Model}                & \textbf{acc.} & \textbf{prec.} & \textbf{recall} & \textbf{f1}     & \textbf{iou}    \\ \hline \hline
 ResNet-50 $ \xrightarrow[]{}$ Fine & 0.810            & 0.774             & 0.684          & 0.680          & 0.550          \\ \hline
InceptionV3  $ \xrightarrow[]{}$  Fine & 0.827            & 0.800             & 0.651          & 0.662          & 0.542          \\ \hline
\textbf{VGG16  $ \xrightarrow[]{}$  Fine}       & \textbf{0.856}   & 0.810    & 0.711 & \textbf{0.711} & \textbf{0.597} \\ \hline
\hline
InceptionV3 $ \xrightarrow[]{}$ None & 0.779 & 0.634   & 0.759 & 0.628 & 0.510 \\ \hline
ResNet-50 $ \xrightarrow[]{}$ None  & 0.769 & 0.632 & 0.777 & 0.647 & 0.516 \\ \hline
VGG-16 $ \xrightarrow[]{}$ None & 0.800 & 0.624 & \textbf{0.839} & 0.666 & 0.551 \\ \hline
\hline
FMASK              & 0.755            & 0.842             & 0.446          & 0.400          & 0.534          \\ \hline
IDEPIX             & 0.736            & \textbf{0.944}    & 0.354          & 0.351          & 0.467           \\ \hline
SEN2COR            & 0.688            & 0.819             & 0.267         & 0.255          & 0.371           \\ \hline
CNN baseline           & 0.737    & 0.718    & 0.643          & 0.617 & 0.494
\\ \hline
\end{tabular}
\end{table}

As a baseline for the proposed method, we consider semantic segmentation algorithms applied on patches, corresponding to the predominant approach to segmentation of large MSI images \cite{shao2019cloud, yang2019cdnet, mateo2017convolutional}. To focus on demonstrating the importance of the proposed multi-scale approach, we consider conventional encoder-decoder CNNs as baselines, matching our fine-grained component. Since the strength of the baseline depends on the choice of encoder backbone and upsampling architecture, we report results for several alternatives.

The set of available CNN architectures is constantly growing, and the set of their possible combinations would become impossibly large for full enumeration. We include a selection of both state-of-the-art and established CNN architectures that we consider representative and limit the combinations as follows. We initially fix the segmentation architecture to UNet  \cite{ronneberger2015u}, applied to cloud segmentation in recent work \cite{jeppesen2019cloud}, while varying the backbone encoders. We consider encoders
Inception-v3 \cite{szegedy2016rethinking}, ResNet-50 \cite{he2016deep}, EfficientNet \cite{tan2019efficientnet}, SEResNeXt-50 \cite{hu2018squeeze}, and VGG-16 \cite{simonyan2014very} in Table~\ref{exp1_table}.
We then vary the segmentation architecture, evaluating
UNet \cite{ronneberger2015u}, Linknet \cite{chaurasia2017linknet}, FPN \cite{lin2017feature}, and PSPNet \cite{zhao2017pyramid}, now fixing the encoder to SEResNeXt-50 that initially performed best with UNet. Table~\ref{exp1_1_table} indicates that in terms of accuracy, best results for the considered encoder-decoder architectures are obtained with a combination of FPN and SEResNext-50. While some variants (UNet/EfficientNet and LinkNet/SEResNext-50) have higher recall, they simultaneously show a significant drop in accuracy.

\subsection{Coarse encoder selection and method validation}
\label{exp1}

For the coarse component to be trained with the undersampled derived dataset of complete images, we considered the same encoder architectures as for the fine-grained model. We modified these for use as the coarse component as described in \ref{ch_coarse_model} but omitted EfficientNet and SEResNeXt-50 due to technical and memory limitations of the test environment. To  demonstrate the effect of prepending our coarse component in a cascade, using otherwise identical choices in the proposed method and a CNN baseline, we assume the FPN/SEResNext-50 as the fine-grained model component, the same network having been evaluated above as a baseline. We additionally report results for a simplified variant that omits the fine-grained model altogether and simply maps all pixels in "Partly Cloudy" patches as "Cloud". This naturally maximizes recall and is included in the results to illustrate the importance of modeling global features.

Table~\ref{exp1_2_table} reports the metrics for the model variants, using the test images. To provide remote sensing context for the results, we report the same metrics also for three well-known MSI cloud masking algorithms (FMask \cite{zhu2015improvement}, Sen2Cor \cite{richter2012sentinel} and Idepix \cite{lebreton2016cloud})\footnote{The metrics for these masks are reported at 20m, their highest available common resolution. Note that the impact of undersampling on the numbers is negligible; e.g. the proposed method evaluated at 20m resolution has $0.8567$ accuracy compared to the $0.8564$ reported for full 10m resolution.}. For the Idepix baseline mask, we include pixels marked as \textit{f\_cloud, f\_cirrus\_sure, f\_cirrus\_ambiguous, f\_clear\_snow, f\_cloud\_shadow or f\_brightwhite}, for maximal accuracy and recall against our dataset. For Sen2Cor, we include \textit{"Cloud Shadows", "Clouds (low to high probability)", "Cirrus" and "Snow/Ice"}.

Irrespective of the choice of the coarse encoder, the proposed method outperforms all of the existing cloud detection models by a wide margin. 
We also confirm high recall compared to the alternatives, matching a key motivation of the work.
Figure~\ref{fig_results} illustrates the difference for a single test image.

For these comparisons, we used the class-wise weighted cross-entropy loss with hyperparameter values of $\alpha = 2$ and $\beta = 0.5$ for the "Partly Cloudy" class and $\alpha = \beta = 1$ for others, setting $\gamma=0$ for all classes. That is, the best result did not use the adjacency loss, which in our experiments nevertheless can better capture the coarse cloud boundary at a marginal cost on overall accuracy, as shown next.

\subsection{Effect of loss function elements}
\label{exp2}

\begin{table}[t]
\caption{Effect of coarse model loss function choice on internal (coarse-acc) and overall (accuracy) performance, as well as the normalized adjacency score measuring how well the borders between classes are captured. We compare variants of our loss function against standard categorical cross-entropy (CCE).}
\label{exp2_table}
\begin{tabular}{|l|l|l|l|}
\hline
\textbf{Loss function}            & \textbf{coarse-acc} & \textbf{accuracy} & $\mathbf{S_{\mathrm{dn}}}$ \\ \hline \hline
CCE & 0.813                          & 0.718                          & 0.09781                            \\ \hline
$\mathcal{L}_{\mathrm{wce}}$              & \textbf{0.855}                 & \textbf{0.812}                 & 0.01778                            \\ \hline
$\mathcal{L}_{\mathrm{adj}} + \mathcal{L}_{\mathrm{wce}}$ & 0.846                          & 0.809                          & \textbf{0.00968}                    \\ \hline
\end{tabular}

\end{table}

The motivation to use a specific loss function  for training the coarse model, stemming from the inherently small proportion of partly cloudy patches,  was analyzed in Section~\ref{lossrationale}. Table \ref{exp2_table} measures the effect of the custom loss function elements described, comparing the proposed class-wise weighted cross entropy (WCE) loss with and without the adjacency penalty against the standard choice of categorical cross entropy (CCE) loss. We evaluate the effect of the loss in context of one example configuration, a coarse ResNet-50 with fine-grained ResNet-50/Unet. For WCE we use the same weighting as in Section~\ref{exp1}, and for the adjacency loss weight, we use $\gamma=1$. Since an adjacency loss $\mathcal{L}_{\mathrm{adj}}$ term is defined for a pair of classes, we use a separate $\mathcal{L}_{\mathrm{adj}}$ term for each class pair out of the three classes, penalizing deviations in the count of "Overcast/Cloudless" pairs with  a weight of $0.75$ and the other two pairs ("Overcast/Partly Cloudy" and "Cloudless/Partly Cloudy") by $0.125$.

The result is that both of the improved loss functions clearly outperform the common CCE in terms of pixel accuracy of the final segmentation (accuracy) as well as the internal accuracy of the coarse model (coarse-acc). Importantly for reducing the unwanted adjacencies of Overcast/Cloudless, the weighted loss functions improve the target metric of normalized adjacency score ($S_{\mathrm{dn}})$ for the class pair roughly by a factor of ten, which indicates that they reproduce the proportions of class boundary regions considerably better. This can be visually verified in Figure~\ref{fig_lossrationale} that shows sample predictions for each loss.
The class-wise weighted cross-entropy in itself already improves boundary detection notably, but interestingly, a further improvement by a factor of two is achieved in discerning the cloud boundary by activating the adjacency term, only negligibly impacting accuracy ($0.809$ vs. $0.812$). This would have made $\mathcal{L}_{\mathrm{adj}} + \mathcal{L}_{\mathrm{wce}}$ an equally valid basis for the previous experiment on overall performance, providing additional smoothness at mask borders.

\subsection{Computation time}

Segmenting clouds on a single high-resolution Sentinel-2 MSI image using the proposed dual model framework, the proposed dual-network architecture ran 4.1 times faster compared to the FPN/SEResNeXt baseline, taking roughly 9s per image vs 37s for the baseline on a NVIDIA GTX10180 GPU. The improvement is explained by a reduction of almost 80\% in the number of patches processed. The time overhead added by the coarse model is negligible, less than 1s per image.

\subsection{Discussion}

From the point of view of the semantic segmentation problem, the proposed framework outperforms the selected CNN baseline (see Table \ref{exp1_2_table}) by all chosen metrics for a cloud mask dataset with characteristics of contiguity and enhanced cloud recall. For instance, pixel accuracy was improved from 73.7\% to 85.6\%. i.e. by a relative accuracy improvement of 16\%. 
In view of the cloud detection application in remote sensing, we achieved a 13\% relative improvement in accuracy to the closest considered cloud detection algorithm. This, alongside validating our method of approach, also suggests that the proposed method has potential to narrow down the gap between applications' needs for sensitive cloud masks vs. available cloud masking methods.

Interestingly, our coarse model architecture reaches 80\% accuracy also when the fine-grained model is not used at all. In this variant, we set all patches marked as "Partly Cloudy" to contain "Cloud" pixels only and as a result, we outperform the CNN baselines based on patching. This highlights the importance of modeling larger spatial features, especially in tasks where the output needs to be largely contiguous, and suggests steering more effort in MSI segmentation towards models inspecting larger proportions of the whole image in contrast to majority of the current literature; see Section~\ref{sec:related}.

Our task of specifically reproducing the given approximate and contiguous annotations, based both on geospatially widespread and pixel-level information, is more challenging than typical supervised CNN setups where target objects have boundaries, or annotations are focused on the pixel. Still, the overall accuracy can be contrasted to those reported in literature for standard masks. In a recent study, Baetens et al. \cite{baetens2019validation} report accuracies in the range of $84-91\%$ for  validating a generated reference annotation against FMask, MAJA and Sen2Cor. We reach accuracy that is within the same range in a more challenging task. This validates that the overall architecture and data pipeline reflect the state of the art in the field. On the other hand, we demonstrated that many existing masks are not sufficient for recreating the annotations in our data and in particular have very poor recall, between 0.27 and 0.45, compared to  0.71 (or 0.84 if mapping the whole border area to "Cloud") of the model trained on these annotations.

Absolute accuracy, for developing the framework into a production application, could be further improved e.g. by 
taking advantage of readily available MSI image QA bands e.g. snow and ice in e.g. preprocessing a subtraction from ground truth mask and treatment as distinct classes, in the style of \cite{mohajerani2018cloud},
or by searching for an optimal fine-grained architecture and the hyperparameters $\alpha, \beta$, and $\gamma$ using a more systematic validation procedure. 

\section{Conclusion}

We propose a novel two-phase semantic segmentation framework for cloud detection from high-resolution optical remote sensing images, drawing on state-of-the-art CNN architectures. We train the components of the CNN framework with cloud masks manually annotated with sensitivity for sparse regions of cloud-ambiguous or hazy pixels. Our dataset, originally collected during construction of cloudless mosaics for a land cover project\cite{corinelc}, contains images from within the growing season, including bare ground, with residual snow, ice and clouds included in a single class of an exclusion mask. To identify image patches that need the most fine-grained cloud detection, and also to reproduce a contiguous quality present in the given annotated masks, we use a modified VGG architecture at an undersampled input resolution. Here, we use a weighted loss function to handle an inherent class imbalance. To classify pixels of the cloud-ambiguous patches at full resolution, we pass them on to a second CNN component based on a SEResNeXt-FPN architecture \cite{hu2018squeeze}\cite{lin2017feature} . The overall framework allows us to analyze large images in a wider variation of scale than otherwise would be possible in the usual setup of a single CNN. Our experiments show a relative accuracy improvement of 16\% by our aggregated dual model over baselines of well-known encoder-decoder CNN architectures trained on image patches. 

As a semantic segmentation framework of large images, the proposed solution is not limited to the domain of remote sensing or cloud detection. We apply the framework to a high-resolution Sentinel-2 dataset but expect it to be readily applicable to semantic segmentation of other multispectral datasets such as Landsat or MODIS and to perform well especially in scenarios of liberally annotated, highly contiguous masks over features having ambiguous boundaries.

\section*{Acknowledgements}
We thank Finnish Environment Institute's Data and Information Centre, particularly Eeva Bruun and Tommi Karesvuori, for annotating and providing the cloud mask dataset, and Jorma Rinkinen of Operos Oy for performance tuning insights regarding our data pipeline. This work was supported by the Academy of Finland Flagship programme: Finnish Center for Artificial Intelligence, FCAI.

\balance

\bibliographystyle{IEEEtran}  
\bibliography{references} 

\end{document}


\title{Supplementary material for: \\ Multi-scale cloud detection in remote sensing images using a dual convolutional neural network}

\author{Markku~Luotamo$^*$,~
        Sari~Metsämäki~
        and~Arto~Klami
\thanks{$^*$M.Luotamo and A.Klami: University of Helsinki, Dept. of Computer Science}%
\thanks{Sari Metsämäki: Finnish Environment Institute}%
}

\markboth{}%
{Luotamo \MakeLowercase{\textit{et al.}}: Multi-scale Cloud Detection in Remote Sensing Images using a Dual Convolutional Neural Network}

\addeditor{AK}
\addeditor{ML}
\addeditor{SM}

\maketitle

\IEEEpeerreviewmaketitle

\section{Introduction}
\label{supp:intro}

This supplementary material provides additional technical details for a manuscript with the same title.
Section~\ref{sec:pipeline} explains the full data processing pipeline and provides justification for the various resolution choices presented in the manuscript. Section~\ref{sec:onehot} presents technical details for the differentiable approximation used for constructing the loss function for the model. Finally, Section~\ref{sec:unet} presents a full specification for one of the neural network architecture baselines used for pixel-level segmentation as the "fine-grained model".

\section{Data Pipeline Dimensions}
\label{sec:pipeline}

\begin{figure*}
  \includegraphics[width=\textwidth, trim=0 0 0 5cm, clip, center]{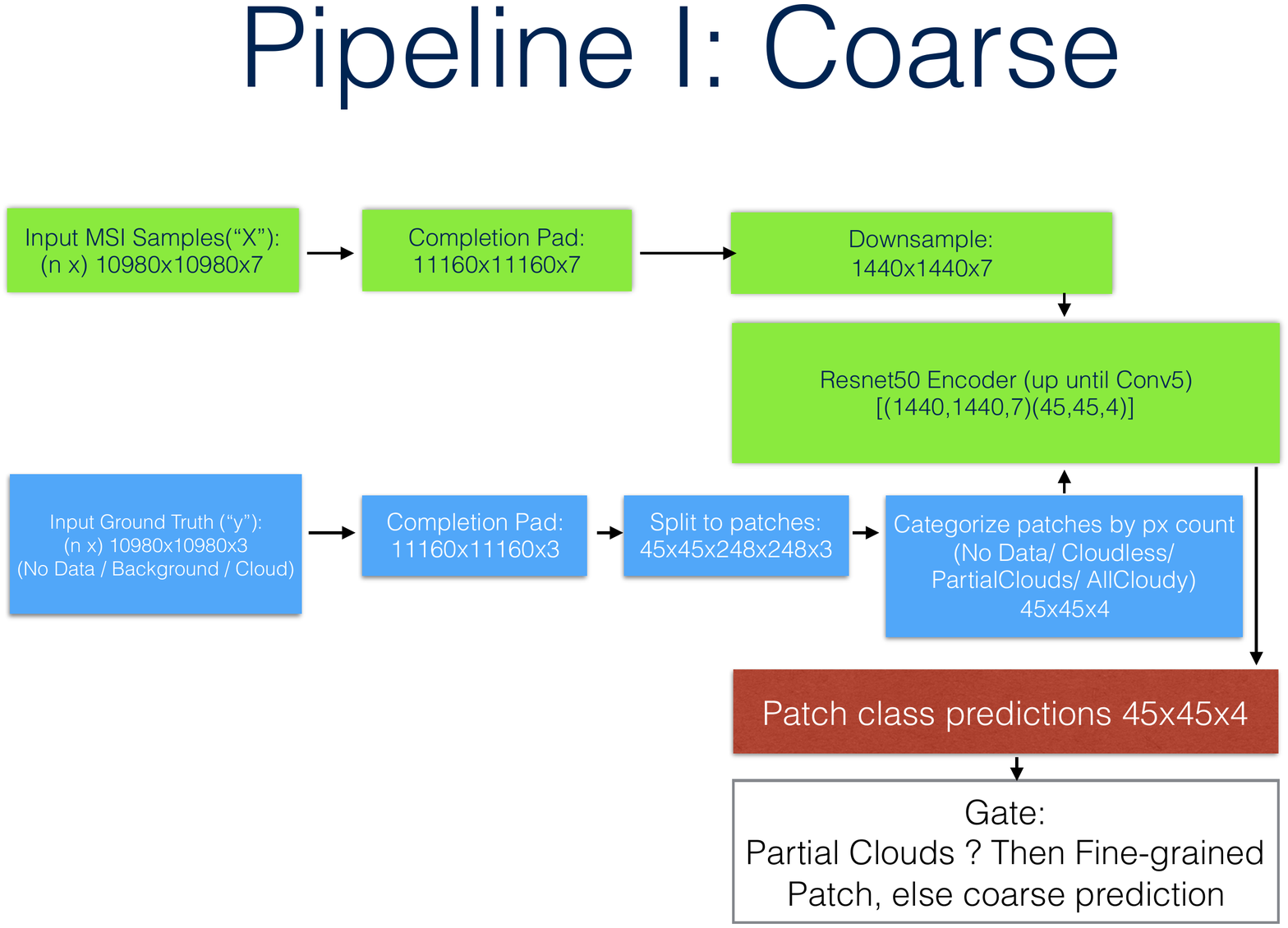} 
  \caption{Data pipeline for the coarse model component.}
  \label{coarse_arch}
\end{figure*}

\begin{figure*}
  \includegraphics[width=\textwidth, trim=0 0 0 5cm, clip]{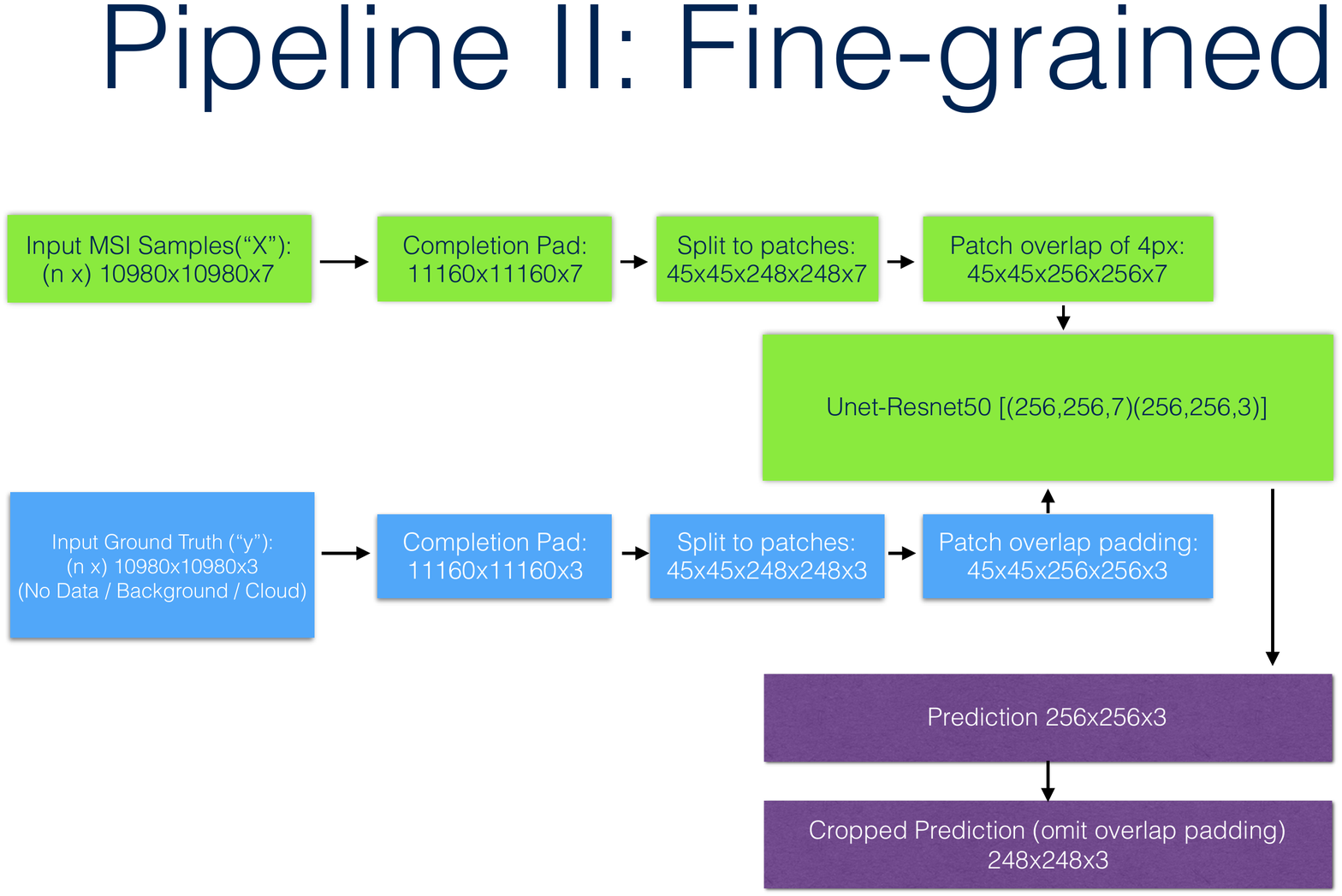}
  \caption{Data pipeline for the fine-grained model component.}
  \label{fineg_arch}
\end{figure*}

The manuscript as such provides a high-level overview of the data pipeline, for brevity skipping rationale and details. 
In this section, we cover the remaining details and calculate the dimensions in a reverse order from end to beginning, from a computationally convenient patch size of the fine-grained model (see Fig. \ref{fineg_arch}) towards the constraint of a Sentinel-2 MSI resolution of 10980x10980. This yields the dimensions of the coarse model used earlier in the processing steps (See Fig. \ref{coarse_arch}). The details are presented for the specific input size, but we note that a similar procedure could easily be carried out for images of other dimensionalities.

For training deep neural networks on a GPU, input image size of a multiple of 8 is commonly used and recommended in the industry (see for instance \cite{nvidiadlperf}). We apply this recommendation with Tensorflow and an NVidia GPU GTX1080 in our use: we select 256x256, an increase from the most common 224x224 patch size to allow slightly larger features from the start. We reserve a 4-pixel margin from the patch for overlap to overcome edge artifacts we observed in early experiments. This results in an effective patch size of 248x248. Dimensions of Sentinel-2 MSI data, 10980x10980 px, are not divisible by 248 (or 256 or 224, for that matter), so the images and the ground truth need to be appended with slices we call a completion pad, to 11160x11160 px in order to be the closest multiple of 248 (45*248=11160). We can now slice the 11160x11160 data to 45x45 patches of 248x248 px effective size,that is, 256x256px with the overlap pixels, ready to be input to the fine-grained model. In the end we hence use 256x256x7 slices of the original image and 256x256x3 masks as data input.

Let us then assume a 45x45 bottleneck resolution for our coarse model. 32:1 is the spatial ratio found in many CNN segmentation encoder backbones between input resolution and the bottleneck layer for height and width. When 45x45 is set as the bottleneck resolution, we get an input resolution of 1440x1440px for the coarse model. On the other hand 1440x1440 approaches the empirical upper limit for 7-band image size for the given GPU memory size and suggested neural network. We downsample the 11160x11160 image to 1440x1440, which provides the coarse model with input.

The coarse model ground truth masks, however, need to be of shape 45x45x4 for classifying each patch of 248x248 on the completion-padded image. These are obtained from the  completion-padded 11160x11160x3 ground truth by slicing it to a grid of 45x45x248x248x3 (no overlap). The 248x248 dimensions are eliminated and the last dimension of 3 classes transformed to 4 by pooling the "cloudy" pixels of a patch to four classes (fully overcast/partly cloudy/cloudless/no data) based on pixel count per original class annotation within a patch. The full-resolution "cloudy" pixel count $n_{pc}$ condition for a ground truth patch to be automatically annotated to "partly cloudy" is $0 < n_{pc} < w \times h$ for our experiments below, where $w$ and $h$ are width and height of the patch, i.e. 248x248.

Altogether, this data pipeline supports conditional two-phase processing based on the result of the coarse classification as explained in the article. Only "partly cloudy" patches on the edges of cloudy regions need to be delegated for more granular classification of the patch, to the fine-grained model.

\section{Differentiable one-hot encoding}
\label{sec:onehot}

Neural networks are trained with gradient-based methods, which implies that the loss function must be  differentiable.
Besides approximating the step function as explained in the manuscript, we replace the one-hot encoding $\mathrm{\sigma}()$ with a soft approximation presented as TensorFlow code in Figure~\ref{fig:listing}.


\begin{figure*}[t]
\begin{code}
\label{code:python-code}
\begin{minted}{python}
def soft_onehot(x):
    return -(tf.sign(tf.reduce_max(x, axis=-1, keepdims=True) - x) - 1)
\end{minted}
\end{code}
\caption{TensorFlow code for a differentiable one-hot function used in the coarse model loss function.}
\label{fig:listing}
\end{figure*}

\begin{figure*}[ht!]
  \includegraphics[width=1.8\textwidth, trim=1cm 5cm 0 1cm, clip, left]{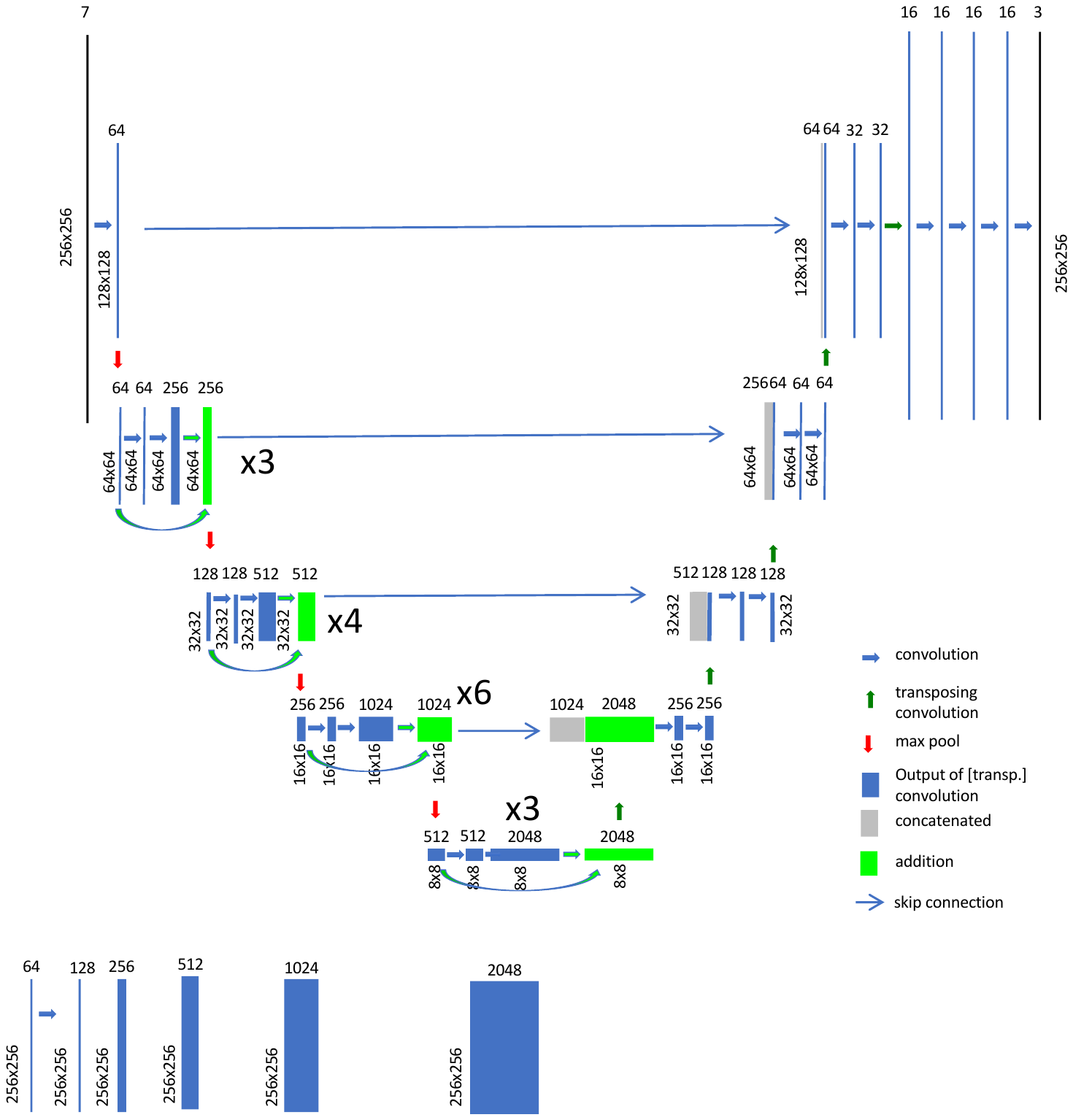} %
  \caption{A possible fine-grained segmentation model structure that draws on U-Net,  with a Resnet-50 encoder backbone}
  \label{fineg_arch_illustr}
\end{figure*}

\section{Fine-grained model architecture}
\label{sec:unet}

In the article, our work evaluated 11 combinations of different CNN architectures, for brevity omitting experiments with less relevant results. As a representative and straightforward example to illustrate the relationship and propagation of the dimensions from the input images to e.g. the bottleneck layer, we show in Figure~\ref{fineg_arch_illustr} a UNet \cite{ronneberger2015u}, with a ResNet \cite{he2016deep} backbone with all of the layer sizes embedded into the figure. The networks are also representative due to their archetypal use of lateral and residual skip connections often present in the evaluated networks from a later date. To make alternative component architectures readily interchangeable, most base implementations were used from a Segmentation models library by 
Yakubovskiy \cite{qubvelsegmodel}. The modifications required by the coarse component included first truncating a backbone network from the end up to the encoder bottleneck layer, then appending 1x1 convolutions and a softmax layer to yield a 45x45 classification as explained with more context in the article. 

\bibliographystyle{IEEEtran}  
\bibliography{references}